\documentclass[10pt,twocolumn,letterpaper]{article}
\usepackage{cvpr}          

\usepackage{graphicx}
\usepackage{amsmath}
\usepackage{multirow}
\usepackage{multicol}
\newtheorem{theorem}{Theorem}
\usepackage{amssymb}
\usepackage{booktabs}
\usepackage[ruled,vlined]{algorithm2e}

\usepackage[pagebackref,breaklinks,colorlinks]{hyperref}

\usepackage[capitalize]{cleveref}
\crefname{section}{Sec.}{Secs.}
\Crefname{section}{Section}{Sections}
\Crefname{table}{Table}{Tables}
\crefname{table}{Tab.}{Tabs.}

\def\cvprPaperID{9439}
\def\confName{CVPR}
\def\confYear{2023}

\begin{document}

\title{CFA: Class-wise Calibrated Fair Adversarial Training}

\author{
    Zeming Wei\textsuperscript{1}, 
    Yifei Wang\textsuperscript{1},
    Yiwen Guo\textsuperscript{2},
    Yisen Wang\textsuperscript{3,4}\thanks{Corresponding Author: Yisen Wang (yisen.wang@pku.edu.cn)}\\
    \textsuperscript{1}School of Mathematical Sciences, Peking University 
    \textsuperscript{2}Independent Researcher\\
    \textsuperscript{3}National Key Lab of General Artificial Intelligence \\ School of Intelligence Science and Technology, Peking University\\
\textsuperscript{4}Institute for Artificial Intelligence, Peking University
}

\maketitle

\begin{abstract}
   Adversarial training has been widely acknowledged as the most effective method to improve the adversarial robustness against adversarial examples for Deep Neural Networks (DNNs). So far, most existing works focus on enhancing the overall model robustness, treating each class equally in both the training and testing phases. Although revealing the disparity in robustness among classes, few works try to make adversarial training fair at the class level without sacrificing overall robustness. In this paper, we are the first to theoretically and empirically investigate the preference of different classes for adversarial configurations, including perturbation margin, regularization, and weight averaging. Motivated by this, we further propose a \textbf{C}lass-wise calibrated \textbf{F}air \textbf{A}dversarial training framework, named CFA, which customizes specific training configurations for each class automatically. Experiments on benchmark datasets demonstrate that our proposed CFA can improve both overall robustness and fairness notably over other state-of-the-art methods. Code is available at \url{https://github.com/PKU-ML/CFA}.
\end{abstract}

\section{Introduction}
Deep Neural Networks (DNNs) have achieved remarkable success in a variety of tasks, but their vulnerability against adversarial examples~\cite{szegedy2013intriguing, goodfellow2014explaining} have caused serious concerns about their application in safety-critical scenarios \cite{chen2015deepdriving,ma2019understanding}. DNNs can be easily fooled by adding small, even imperceptible perturbations to the natural examples. To address this issue, numerous defense approaches have been proposed~\cite{papernot2016distillation,das2017keeping,xie2019feature,bai2019hilbert,mo2022adversarial}, among which Adversarial Training (AT)~\cite{madry2017towards,wang2021convergence} has been demonstrated as the most effective method to improve the model robustness against such attacks~\cite{athalye2018obfuscated,wu2020adversarial}.
Adversarial training can be formulated as the following min-max optimization problem:
\begin{equation}
\label{AT}
    \min_{\boldsymbol\theta} \mathbb{E}_{(x, y)\sim\mathcal D} \max_{\|{x'}-{x}\|
    \le \epsilon} \mathcal L({\boldsymbol\theta};{x'},y),
\end{equation}
where $\mathcal D$ is the data distribution, $\epsilon$ is the margin of perturbation and $\mathcal L$ is the loss function, \textit{e.g.} the cross-entropy loss. Generally, Projected Gradient Descent (PGD) attack~\cite{madry2017towards} has shown satisfactory effectiveness to find adversarial examples in the perturbation bound $\mathcal B(x,\epsilon)=\{x':\|x'-x\|\le\epsilon\}$, which is commonly used in solving the inner maximization problem in \eqref{AT}:
\begin{equation}
    x^{t+1}=\Pi_{\mathcal B(x,\epsilon)} (x^t+\alpha\cdot\text{sign}(\nabla_{x^t} \mathcal L(\boldsymbol\theta;x^t,y))),
\end{equation}
where $\Pi$ is the projection function and $\alpha$ controls the step size of gradient ascent. TRADES~\cite{zhang2019theoretically} is another variant of AT, which adds a regularization term to adjust the trade-off between robustness and accuracy \cite{tsipras2018robustness,wang2023simple}:
\begin{equation}
\label{TRADES}
    \min_{\boldsymbol\theta}\mathbb{E}_{(x, y)\sim\mathcal D}\ \{\mathcal L({\boldsymbol\theta};{x},y) + \beta\max_{\|{x'}-{x}\|\le\epsilon} \mathcal{K} (f_{\boldsymbol\theta}({x}), f_{\boldsymbol\theta}({x'}))\},
\end{equation}
where $\mathcal K(\cdot)$ is the KL divergence and $\beta$ is the \textit{robustness regularization} to adjust the robustness-accuracy trade-off. 

Although certain robustness has been achieved by AT and its variants, there still exists a stark difference among class-wise robustness in adversarially trained models, \textit{i.e.}, the model may exhibit strong robustness on some classes while it can be highly vulnerable on others, as firstly revealed in \cite{xu2021robust,tian2021analysis,benz2021robustness}. This disparity raises the issue of robustness fairness, which can lead to further safety concerns of DNNs, as the models that exhibit good overall robustness may be easily fooled on some specific classes, \textit{e.g.}, the stop sign in automatic driving. To address this issue, Fair Robust Learning (FRL)~\cite{xu2021robust} has been proposed, which adjusts the margin and weight among classes when fairness constraints are violated. However, this approach only brings limited improvement on robust fairness while causing a drop on overall robustness. 

In this paper, we first present some theoretical insights on how different adversarial configurations impact class-wise robustness, and reveal that strong attacks can be detrimental to the \textit{hard} classes (classes that have lower clean accuracy). This finding is further empirically confirmed through evaluations of models trained under various adversarial configurations. Additionally, we observe that the worst robustness among classes fluctuates significantly between different epochs during the training process. It indicates that simply selecting the checkpoint with the best overall robustness like the previous method \cite{rice2020overfitting} may result in poor robust fairness, \textit{i.e.}, the worst class robustness may be extremely low.

Inspired by these observations, we propose to dynamically customize different training configurations for each class. Note that unlike existing instance-wise customized methods that aim to enhance overall robustness ~\cite{ding2018mma,balaji2019instance,wang2019improving,cheng2020cat,zhang2020attacks}, we also focus on the fairness of class-wise robustness. Furthermore, we modify the weight averaging technique to address the fluctuation issue during the training process. Overall, we name the proposed framework as \textbf{C}lass-wise calibrated \textbf{F}air \textbf{A}dversarial training (CFA). 

Our contributions can be summarized as follows:
\begin{itemize}
    \item We show both theoretically and empirically that different classes require appropriate training configurations. In addition, we reveal the fluctuating effect of the worst class robustness during adversarial training, which indicates that selecting the model with the best overall robustness may result in poor robust fairness.
    
    \item We propose a novel approach called Class-wise calibrated Fair Adversarial training (CFA), which dynamically
    customizes adversarial configurations for different classes during the training phase, and modifies the weight averaging technique to improve and stabilize the worst class robustness.
    
    \item Experiments on benchmark datasets demonstrate that our CFA outperforms state-of-the-art methods in terms of both overall robustness and fairness, and can be also easily incorporated into other adversarial training approaches to further improve their performance.
    
\end{itemize}

\section{Theoretical Class-wise Robustness Analysis}
In this section, we present our theoretical insights on the influence of different adversarial configurations on class-wise robustness. 

\subsection{Notations}
For a $K$-classification task, we use $f:\mathcal X\to\mathcal Y$ to denote the classification function which maps from the input space $\mathcal X$ to the output labels $\mathcal Y=\{1,2,\cdots,K\}$. For an example $x\in\mathcal X$, we use $\mathcal B(x,\epsilon) = \{x'|\|x'-x\|\le \epsilon\}$ to restrict the perturbation. 
In this paper, we mainly focus on the $l_\infty$ norm $\|\cdot\|_\infty$, and note that our analysis and  approach can be generalized to other norms similarly.

We use $\mathcal A(f)$ and $\mathcal R(f)$ to denote the clean and robust accuracy of the trained model $f$:
\begin{equation}
\begin{split}
    &\mathcal{A}(f) = \mathbb E_{( x,y)\sim\mathcal D}\ \mathbf{1}(f(x) = y ),\\
        &\mathcal{R}(f) = \mathbb E_{( x, y)\sim\mathcal D}\ \mathbf{1}(\forall x'\in\mathcal B(x,\epsilon), f(x')=y).
\end{split}
\end{equation}
We use $\mathcal A_k(f)$ and $\mathcal R_k(f)$ to denote the clean and robust accuracy of the $k$-th class respectively to analyze the class-wise robustness. 

\subsection{A Binary Classification Task}
We consider a simple binary classification task that is similar to the data model used in~\cite{tsipras2018robustness}, but the properties (hard or easy) of the two classes are different. 

\noindent \textbf{Data Distribution.}
Consider a binary classification task where the data distribution $\mathcal D$ consists of input-label pairs $(x,y)\in\mathbb R^{d+1}\times \{-1,+1\}$.
The label $y$ is uniformly sampled, \textit{i.e.,} $y\overset{\text{u.a.r.}}{\sim}\{-1,+1\}$.
For input $x=(x_1,x_2, \cdots, x_{d+1})$, let $x_1\in\{-1,+1\}$ be the \textit{robust feature}, and $x_2,\cdots, x_{d+1}$ be the \textit{non-robust features}. The robust feature $x_1$ is labeled as $x_1 = y$ with probability $p$ and $x_1=-y$ with probability $1-p$ where $0.5 \le p <1$. For the non-robust features, they are sampled from $x_2,\cdots,x_{d+1}\overset{\text{i.i.d}}{\sim}\mathcal N(\eta y, 1)$ where $\eta<1/2$ is a small positive number.
Intuitively, as discussed  in~\cite{tsipras2018robustness}, $x_1$ is robust to perturbation but not perfect (as $p<1$), and $x_2,\cdots, x_{d+1}$ are useful for classification but sensitive to small perturbation. In our model, we introduce some differences between the two classes by letting the probability of $x_1=y$ correlate with its label $y$.
Overall, the data distribution is 
\begin{equation}
    x_1 = \begin{cases}
    + y,& \text{w.p.}\ p_y\\
    - y,& \text{w.p.}\ 1-p_{y}
    \end{cases}  \text{and} \quad
    x_2, \cdots, x_{d+1}\overset{\text{i.i.d}}{\sim}\mathcal N(\eta y, 1).
\end{equation}
We set $p_{+1}>p_{-1}$ in our model.
Therefore, the robust feature $x_1$ is more reliable for class $y=+1$, while for class $y=-1$, the robust feature $x_1$ is noisier and their classification depends more on the non-robust features $x_2, \cdots, x_{d+1}$.

\noindent \textbf{Hypothesis Space.} Consider a SVM classifier (without bias term) $f(x) = \text{sign}(w_1x_1+w_2x_2+\cdots+ w_{d+1}x_{d+1})$. 
For the sake of simplicity,
we assume  $w_1,w_2\ne 0$, and $w_2=w_3=\cdots =w_{d+1}$ since $x_2,\cdots, x_{d+1}$ are equivalent. Then, let $w = \frac{w_1}{w_2}$, the model can be simplified as $f_w(x) = \text{sign}(x_1+\frac{x_2+ \cdots +x_{d+1}}{w})$.
 Without loss of generality, we further assume $w>0$ since $x_2,\cdots,x_{d+1}\sim\mathcal N(\eta y, 1)$ tend to share the same sign symbol with $y$.

\subsection{Theoretical Insights}
\label{sec:theoretical insight}
\begin{figure}[!t]
    \centering
    \begin{tabular}{cc}
        \includegraphics[width=0.22\textwidth]{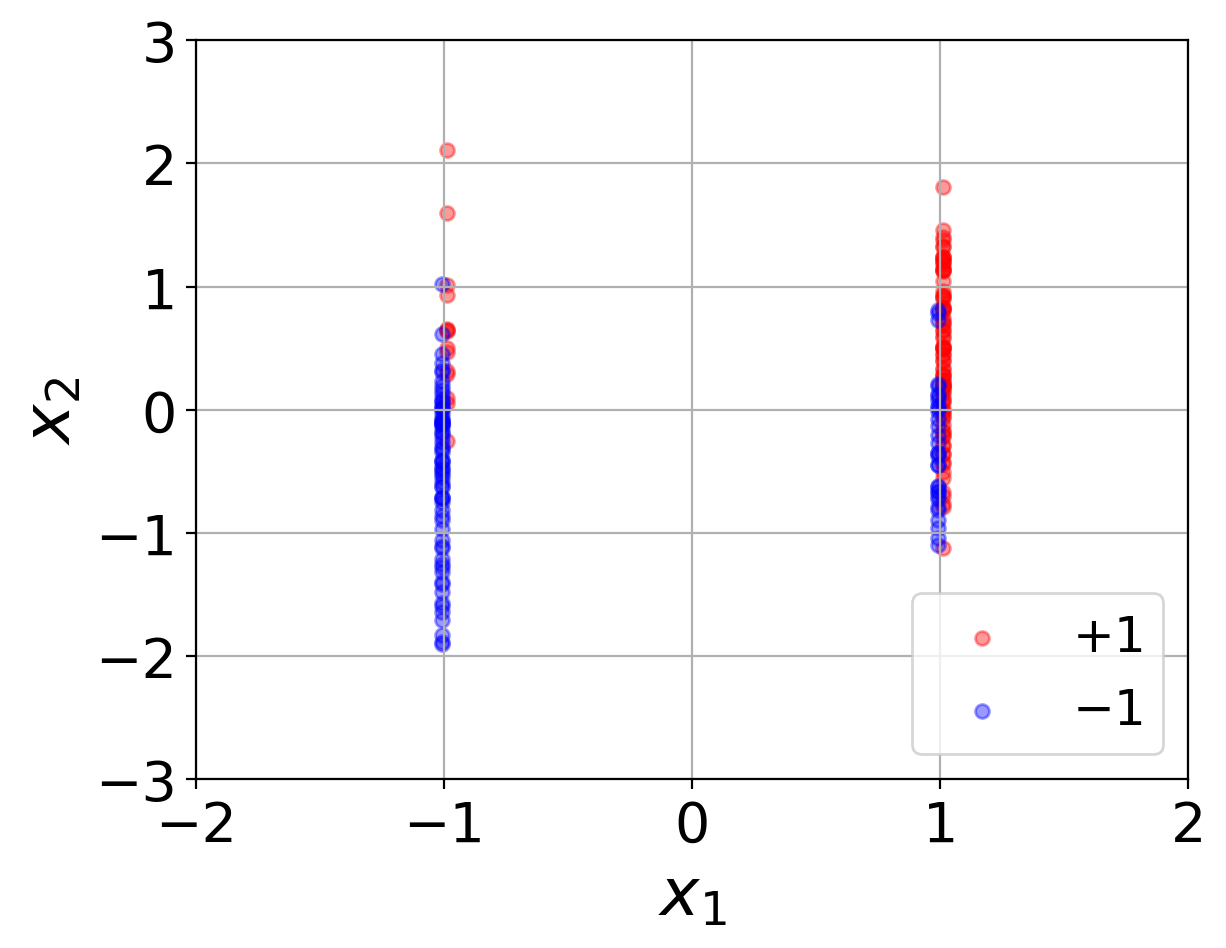} &
        \includegraphics[width=0.22\textwidth]{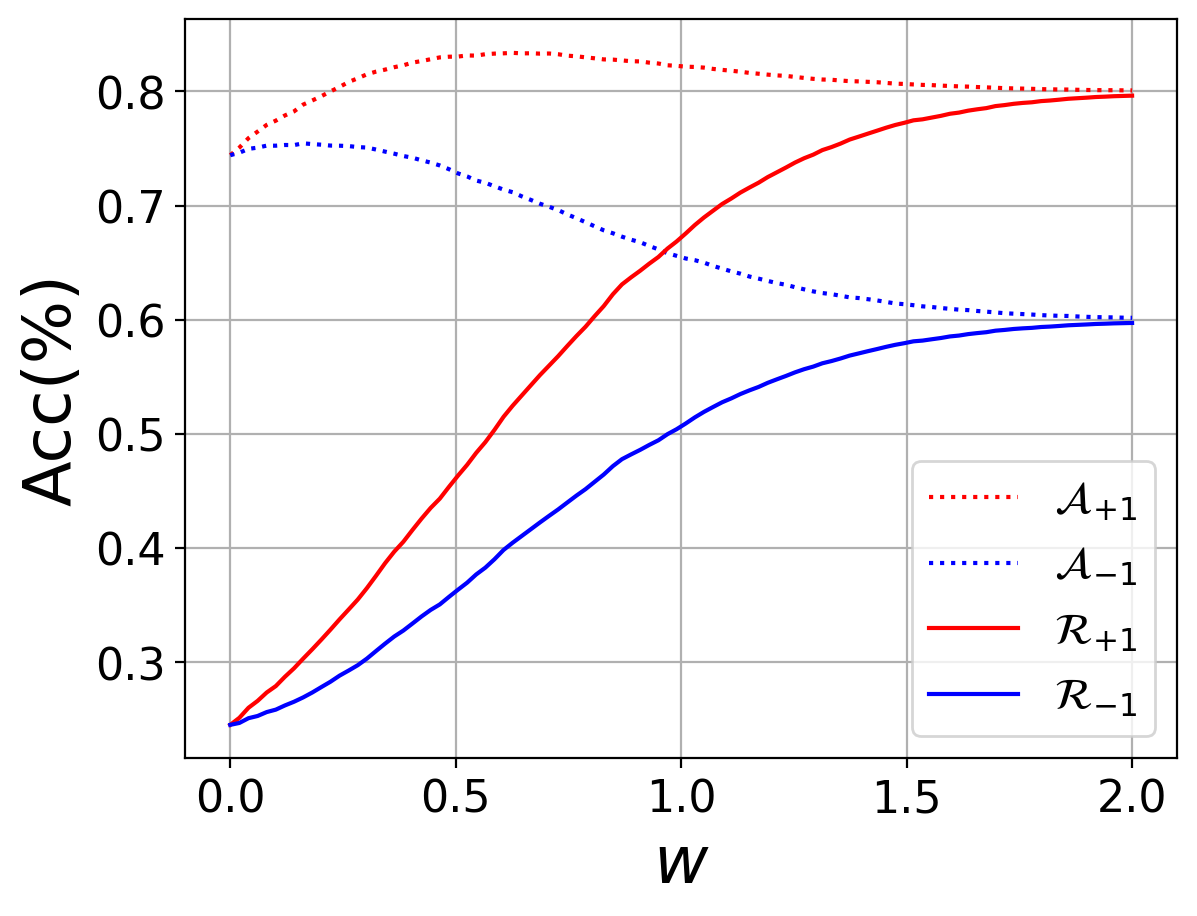}
        \\
        (a) & (b) 
    \end{tabular}
    \vspace{-0.15 in}
    \caption{An visualization of the toy model for the case $d=1$. (a): Sampled data from the distribution. Red dots are labeled $y=+1$ and blue dots are labeled $y=-1$. (b): Clean and robust accuracy of the two classes. Solid lines indicate robust accuracy and dotted lines indicate clean accuracy.}
    \label{fig:toy model}
\end{figure}

\noindent \textbf{Illustration Example.}
An example of the data distribution for the case  $d=1$ is visualized in Fig.~\ref{fig:toy model}(a). The data points for class $y=+1$ are colored red and for $y=-1$ are colored blue. We can see that the robust feature $x_1$ of class $y=-1$ seems to be noisier than $y=+1$, since the frequency of blue dots appearing on the line $x_1=1$ is higher compared to the frequency of red dots appearing on the line $x_1=-1$, with
$p_{+1}>p_{-1}$. Therefore, class $y=-1$ might be more difficult to learn. Furthermore, we plot the clean and robust accuracy of the two classes of $f_w$ for different $w$ in Fig.~\ref{fig:toy model}(b).
Implementation details of this example can be found in Appendix~\ref{illustration Example}. The parameter $w$ can be regarded as the strength of adversarial attack in adversarial training, since larger $w$ indicates the classifier $f_w$ bias less weight on non-robust features $w_2,\cdots,w_{d+1}$ and pay more attention on robust feature $w_1$. 
We can see that as $w$ increases, the clean accuracy of $y=-1$ drops  significantly faster than $y=+1$, but the robustness improves slower. We formally prove this observation in the following. \\

\noindent \textbf{The Intrinsically Hard Class.}
First we formally distinct class $y=-1,+1$ as the \textit{hard} and \noindent \textit{easy} class in Theorem~\ref{theorem:difficulty}.
\begin{theorem} 
For any $w>0$ and the classifier $f_w=\text{sign}(x_1+\frac{x_2+\cdots+x_{d+1}}{w})$, we have $\mathcal A_{+1}(f_w) > \mathcal A_{-1} (f_w)$ and  $\mathcal R_{+1}(f_w)>R_{-1}(f_w)$.
\label{theorem:difficulty}
\end{theorem}
Theorem~\ref{theorem:difficulty} shows that the class $y=-1$ is more difficult to learn than class $y=+1$ both in robust and clean settings. This reveals the potential reason why some classes are intrinsically difficult to learn in the adversarial setting, that is, their robust features are less reliable.\\

\noindent \textbf{Relation Between $w$ and Attack Strength.}
Consider the model is adversarially trained with perturbation margin $\epsilon$. The following Theorem~\ref{theorem:eps} shows using larger $\epsilon$ enlarges $w$.
\begin{theorem} 
\label{theorem:eps}
    For any $0\le\epsilon\le\eta$, let $w^* = \arg\max\limits_w \mathcal R(f_w)$ be
    the optimal parameter for adversarial training with perturbation bound $\epsilon$, then $w^*$  is monotone increasing at $\epsilon$.
\end{theorem}
Theorem~\ref{theorem:eps} bridges the gap between model parameter and attack strength in adversarial training. Next, we can implicitly investigate the influence of attack strength on class-wise robustness by analyzing the parameter $w$. \\

\noindent \textbf{Impact of Attack Strength on Class-wise Robustness.}
Here, we demonstrate how adversarial strength influences class-wise clean and robust accuracy. 

\begin{theorem}
Let $w_y^* = \arg\max\limits_w \mathcal A_y(f_w)$ be the parameter for the best clean accuracy of class $y$, then $w_{+1}^*>w_{-1}^*$.
\label{theorem:best}
\end{theorem}
Theorem~\ref{theorem:best} shows that the clean accuracy of the hard class $y=-1$ reaches its best performance \textit{earlier} than $y=+1$. In other words, $\mathcal A_{-1}(f_w)$ starts dropping earlier than $\mathcal A_{+1}(f_w)$. As the model further distracts its attention from its clean accuracy to robustness by increasing the parameter $w$, the hard class $y=-1$ losses more clean accuracy yet gains less robust accuracy as shown in Theorem~\ref{theorem:compare}.

\begin{theorem} Suppose $\Delta_w>0$, then for $\forall w>w_{+1}^*
$, $\mathcal A_{-1}(f_{w+\Delta_w}) - \mathcal A_{-1}(f_w) < \mathcal A_{+1}(f_{w+\Delta_w}) - \mathcal A_{+1}(f_w)<0$, and for $\forall w>0$, $0<\mathcal R_{-1}(f_{w+\Delta_w}) - \mathcal R_{-1}(f_w) < \mathcal R_{+1}(f_{w+\Delta_w}) - \mathcal R_{+1}(f_w)$. 
\label{theorem:compare}
\end{theorem}

The proof of the theorems can be found in Appendix~\ref{proof}.
In this section, we demonstrate the unreliability of robust features is
a possible explanation for the intrinsic difficulty in learning some classes.
Then, by implicitly expressing the attack strength with parameter $w$, we analyze how adversarial configuration influence class-wise robustness. Theorems~\ref{theorem:best} and \ref{theorem:compare} highlight the negative impact of strong attack on the hard class $y=-1$.

\section{Observations on Class-wise Robustness}
In this section, we present our empirical observations on the class-wise robustness of models adversarially trained under different configurations. Taking vanilla AT \cite{madry2017towards} and TRADES \cite{zhang2019theoretically} as examples, we compare two key factors in the training configurations: the perturbation margin $\epsilon$ in vanilla AT and the regularization $\beta$ in TRADES.
We also reveal the fluctuation effect of the worst class robustness during the training process, which has a significant impact on the robust fairness in adversarial training.

\subsection{Different Margins}
\label{margin analysis}

Following the vanilla AT~\cite{madry2017towards}, we train 8 models on the CIFAR10 dataset~\cite{krizhevsky2009learning} with $l_\infty$-norm perturbation margin $\epsilon$ from $2/255$ to $16/255$ and analyze their overall and class-wise robustness. 
\begin{figure*}[t]
    \centering
    \begin{tabular}{ccc}
    \includegraphics[width=0.3\textwidth]{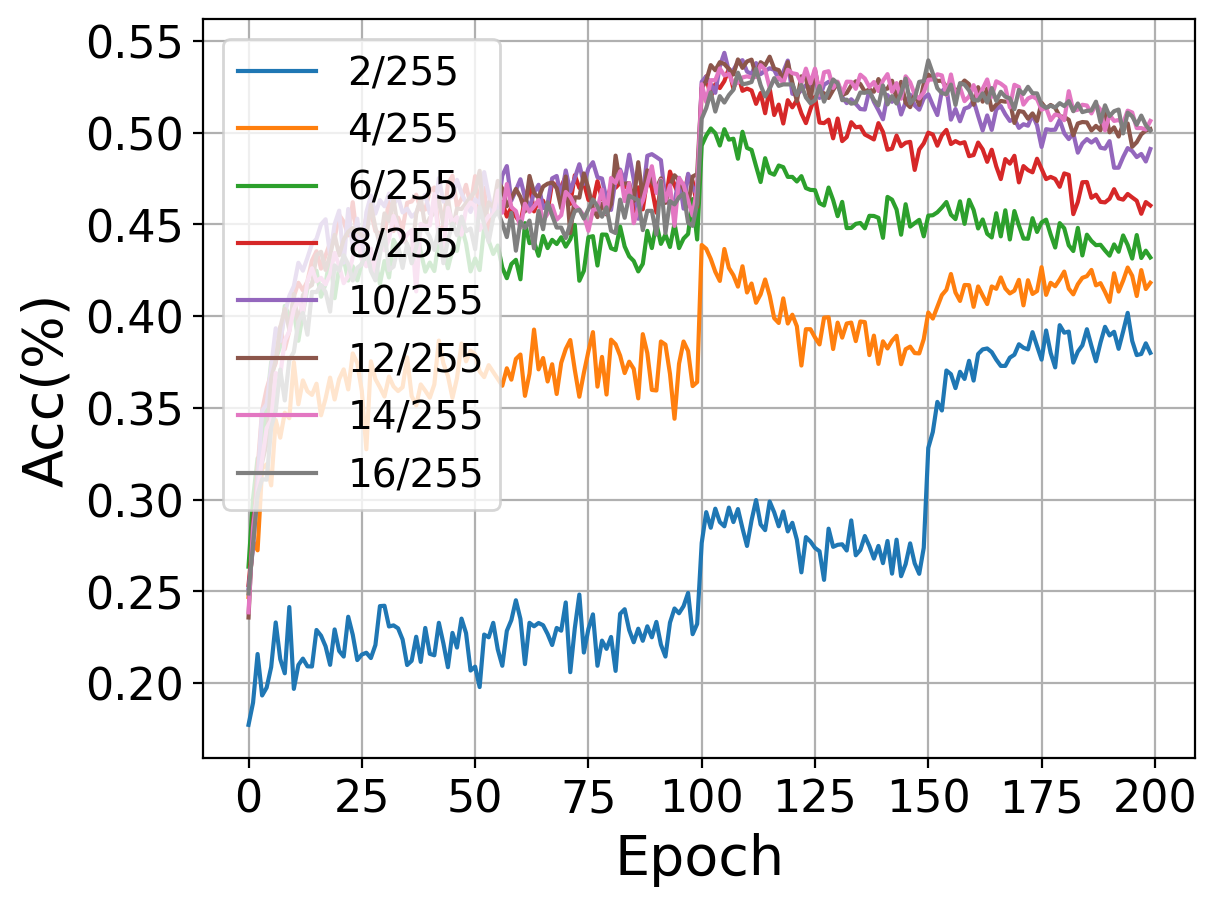} &
    \includegraphics[width=0.3\textwidth]{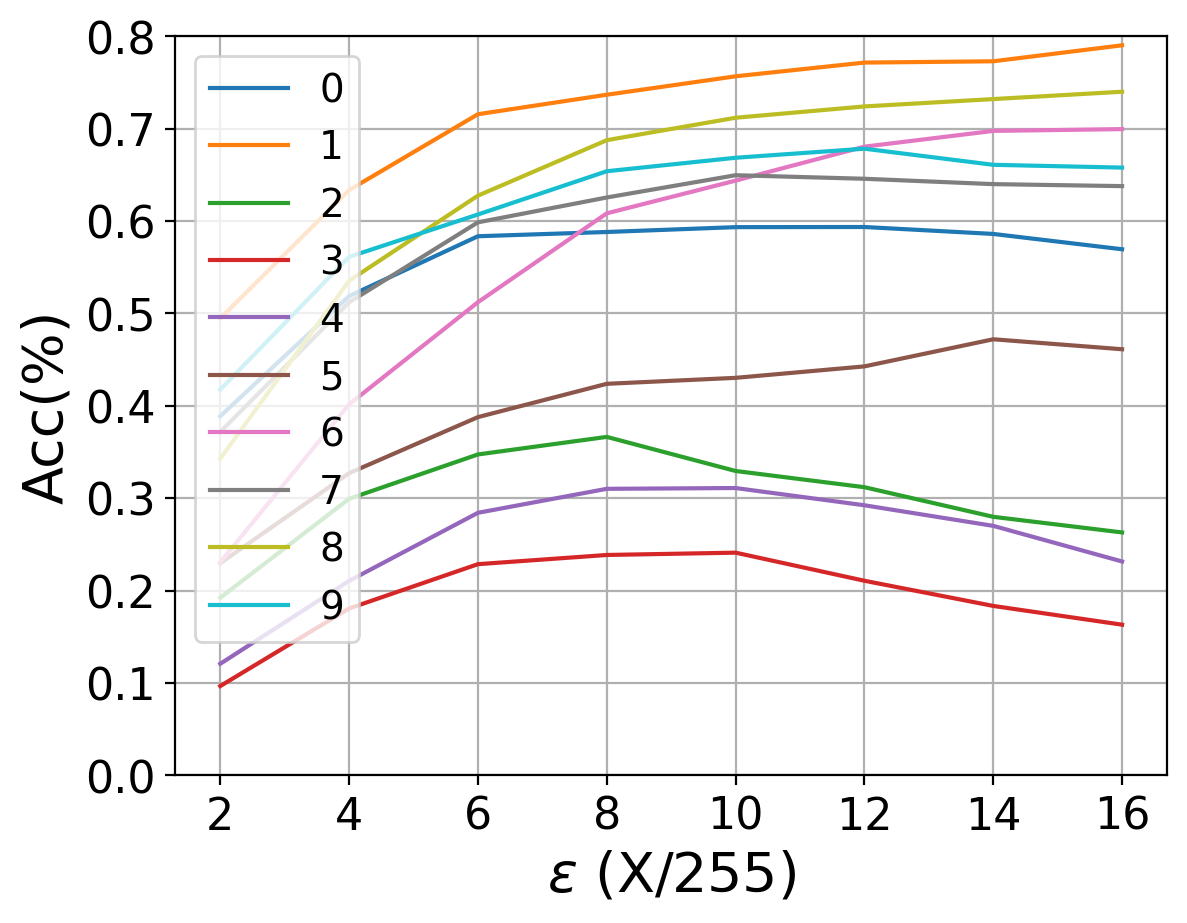} &
    \includegraphics[width=0.3\textwidth]{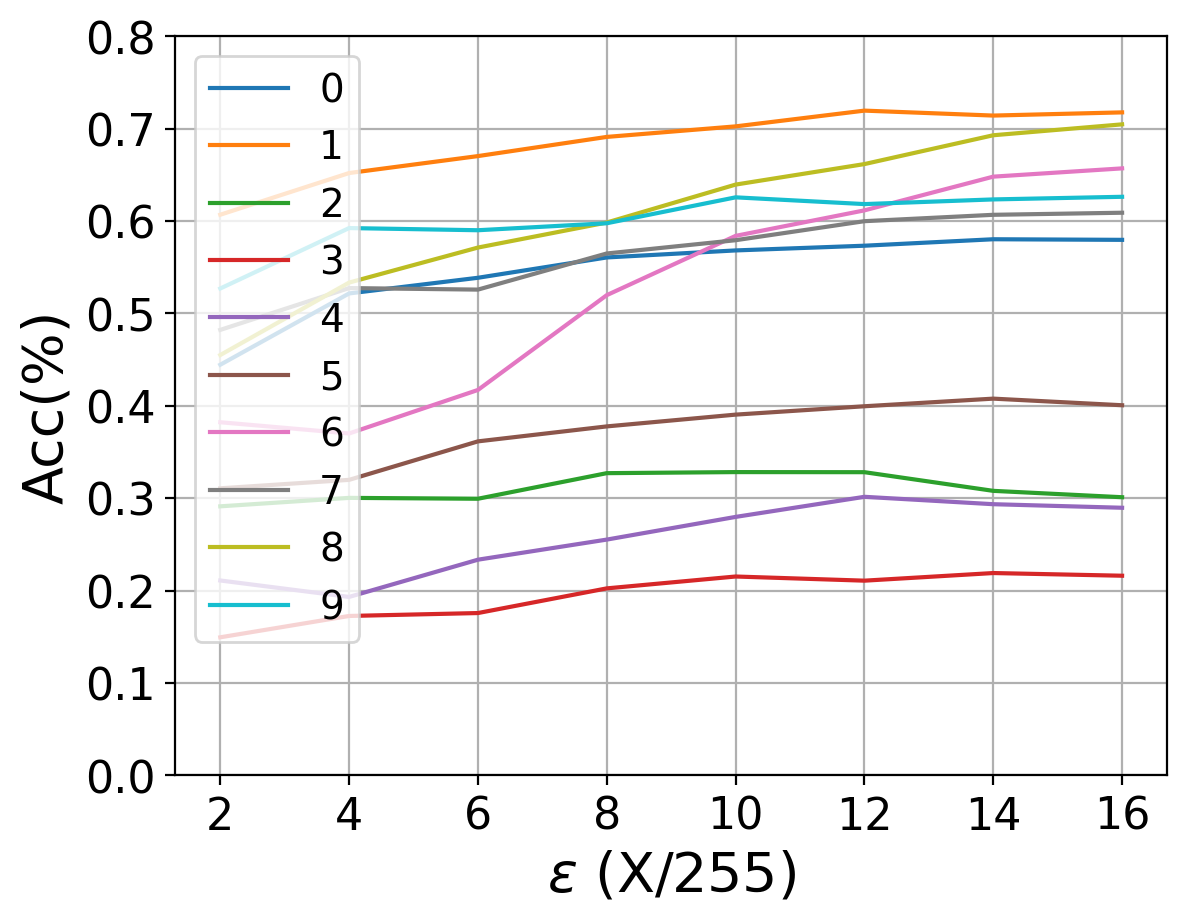}
    \\
    (a) &  
    (b) & 
    (c)
    \end{tabular}
    \vspace{-0.15 in}
    \caption{Comparison of overall and class-wise robustness of models adversarially trained on CIFAR10 with different perturbation margin $\epsilon$. (a): Overall robust accuracy with different perturbation margin $\epsilon$ from 2/255 to 16/255. (b): Average class-wise robust accuracy at epoch $101-120$ (each line represents a class). (c): Average class-wise robust accuracy at epoch $181-200$ (each line represents a class).}
    \label{fig:margin}
\end{figure*}

The comparison of overall robustness is shown in Fig.~\ref{fig:margin}(a). The robustness is evaluated under PGD-10 attack bounded by $\epsilon_0=8/255$, which is commonly used for robustness evaluation.
Intuitively, using a larger margin can lead to better robustness. For $\epsilon<\epsilon_0$, the attack is too weak and hence the robust accuracy of the trained model is not comparable with $\epsilon\ge \epsilon_0$. 
However, for the three models trained with $\epsilon>\epsilon_0$, although their robustness outperforms the case of $\epsilon=\epsilon_0$ at the last epoch, they do not make significant progress on the best-case robustness (around 100-th epoch). 

We take a closer look at this phenomenon by investigating their class-wise robustness in Fig.~\ref{fig:margin}(b) and Fig.~\ref{fig:margin}(c). For each class, we calculate the average class-wise robust accuracy among the 101$-$120-th epochs (where the model performs the best robustness) and 181$-$200-th epochs, respectively. From Fig.~\ref{fig:margin}(b), we can see that a larger training margin $\epsilon$ does not necessarily result in better class-wise robustness. For the \textit{easy} classes which perform higher robustness, their robustness monotonously increase as $\epsilon$  enlarges from $2/255$ to $16/255$. By contrast, for the \textit{hard} classes (especially class $2, 3, 4$), their robustness drop when $\epsilon$ enlarges from $8/255$. However, for the last several checkpoints in Fig.~\ref{fig:margin}(c), we can see a consistent increase on class-wise robustness when the $\epsilon$ enlarges. Revisiting the overall robustness, we can summarize that the class-wise robustness is boosted mainly by reducing the robust over-fitting problem in the last checkpoint.
This can explain why Fair Robust Learning (FRL)~\cite{xu2021robust} can improve robust fairness  by enlarging the margin for the hard classes, since the model reduces the over-fitting problem on these classes. Considering the overall robustness is lower in the last checkpoint (robust fairness is better though), we hope to improve the best-case robust fairness in the situation of a relatively high overall robustness. 

In summary, larger perturbation is harmful to the hard classes in the best case, while can marginally improve the class-wise robustness in the later stage of training. For easy classes, larger perturbation is useful at whatever the best and last checkpoints. Therefore, a specific and proper perturbation margin is needed for each class. 

\subsection{Different Regularizations}
\label{regularizations}
\begin{figure}[h]
    \centering
    \begin{tabular}{cc}
         \includegraphics[width=0.22\textwidth]{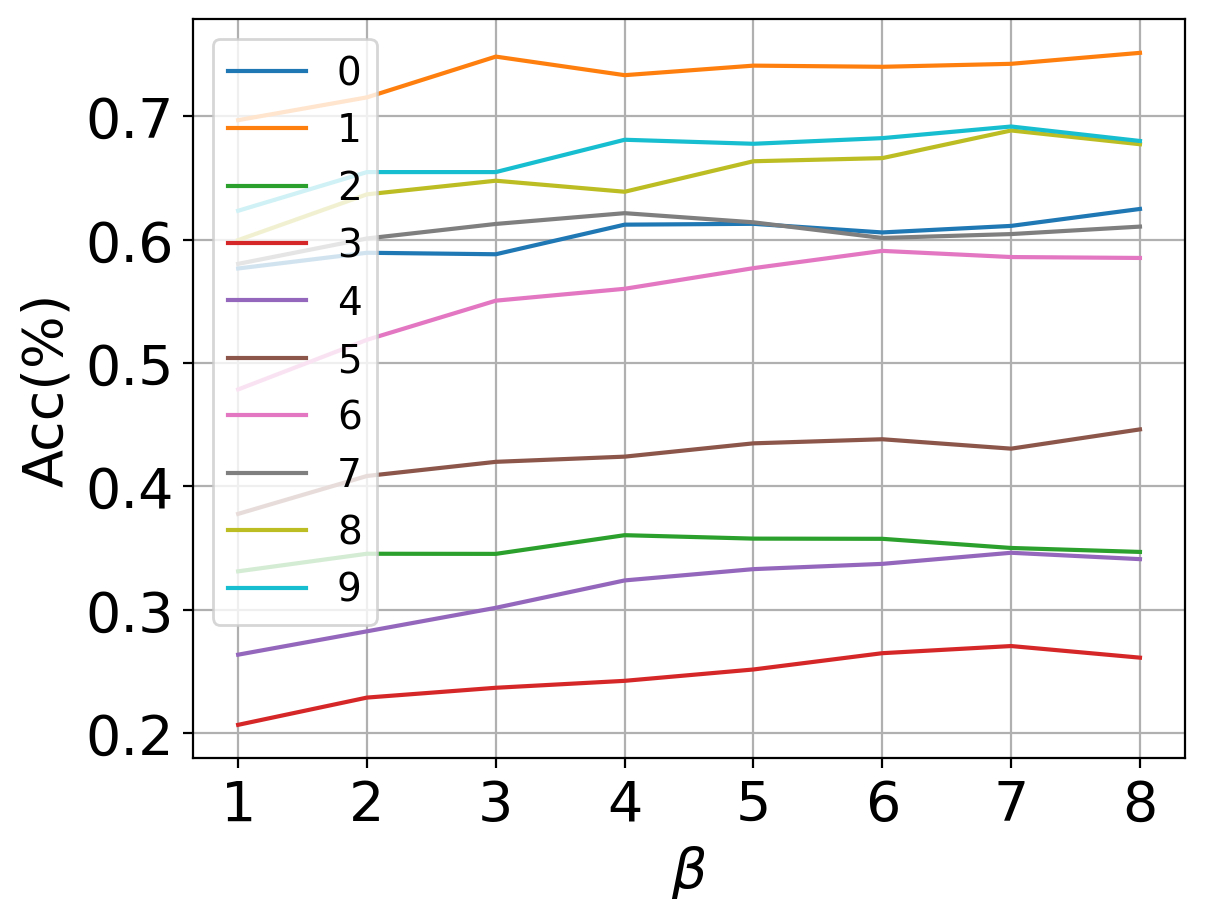}
         & 
         \includegraphics[width=0.22\textwidth]{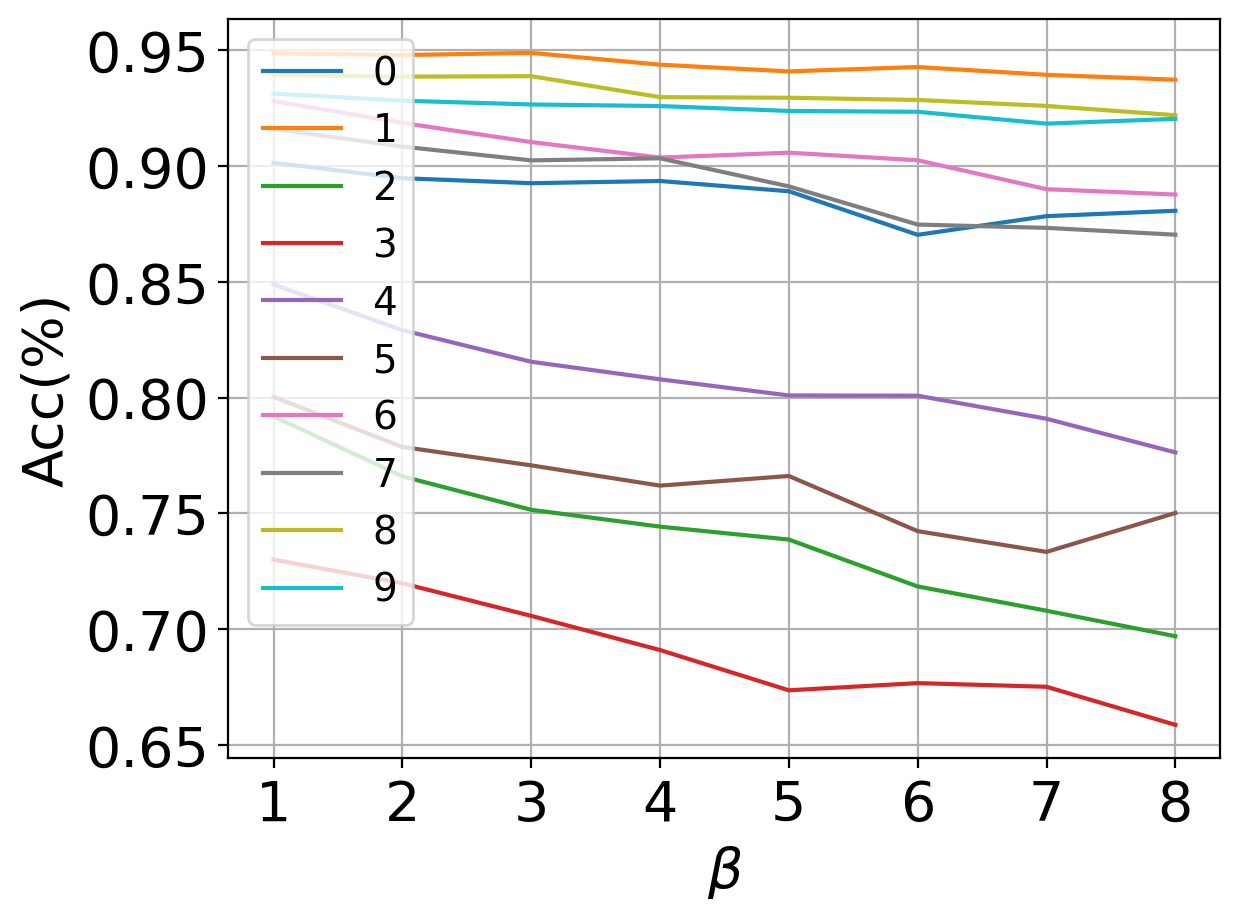}
         \\
         (a) & (b)
    \end{tabular}
    \vspace{-0.15 in}
    \caption{Comparison of class-wise robustness trained by TRADES with different robustness regularization parameters $\beta$. (a) Class-wise robust accuracy. (b) Class-wise clean accuracy.}
    \label{fig:trades}
\end{figure}

In this section, we also conduct a similar experiment on the selection of \textit{robustness regularization} $\beta$ in TRADES. We compare models trained on CIFAR10 with $\beta$ from 1 to 8, and plot the average class-wise robust and clean accuracy among the $151-170$-th epochs (where TRADES performs the best performance) in Fig.~\ref{fig:trades}. We can see that bias more weight on robustness (use larger $\beta$) cause different influences among classes. Specifically, for \textit{easy classes}, improving $\beta$ can improve their robustness at the cost of little clean accuracy reduction, while for \textit{hard classes} (\textit{e.g.}, classes $2,3,4$), improving $\beta$ can only obtain limited robustness improvement but drop clean accuracy significantly. 

\begin{figure}[!htbp]
    \centering
    \begin{tabular}{cc}
        \includegraphics[width=0.22\textwidth]{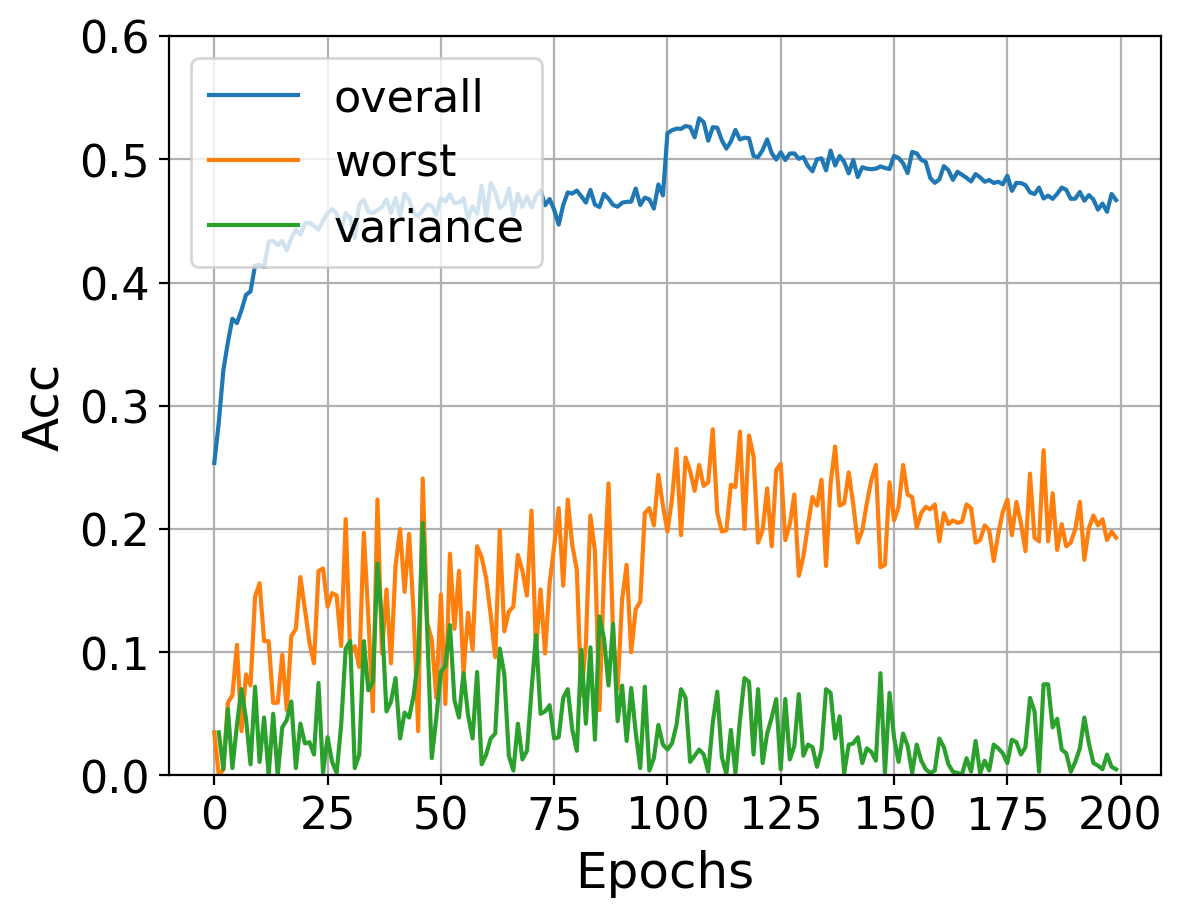} &  
        \includegraphics[width=0.22\textwidth]{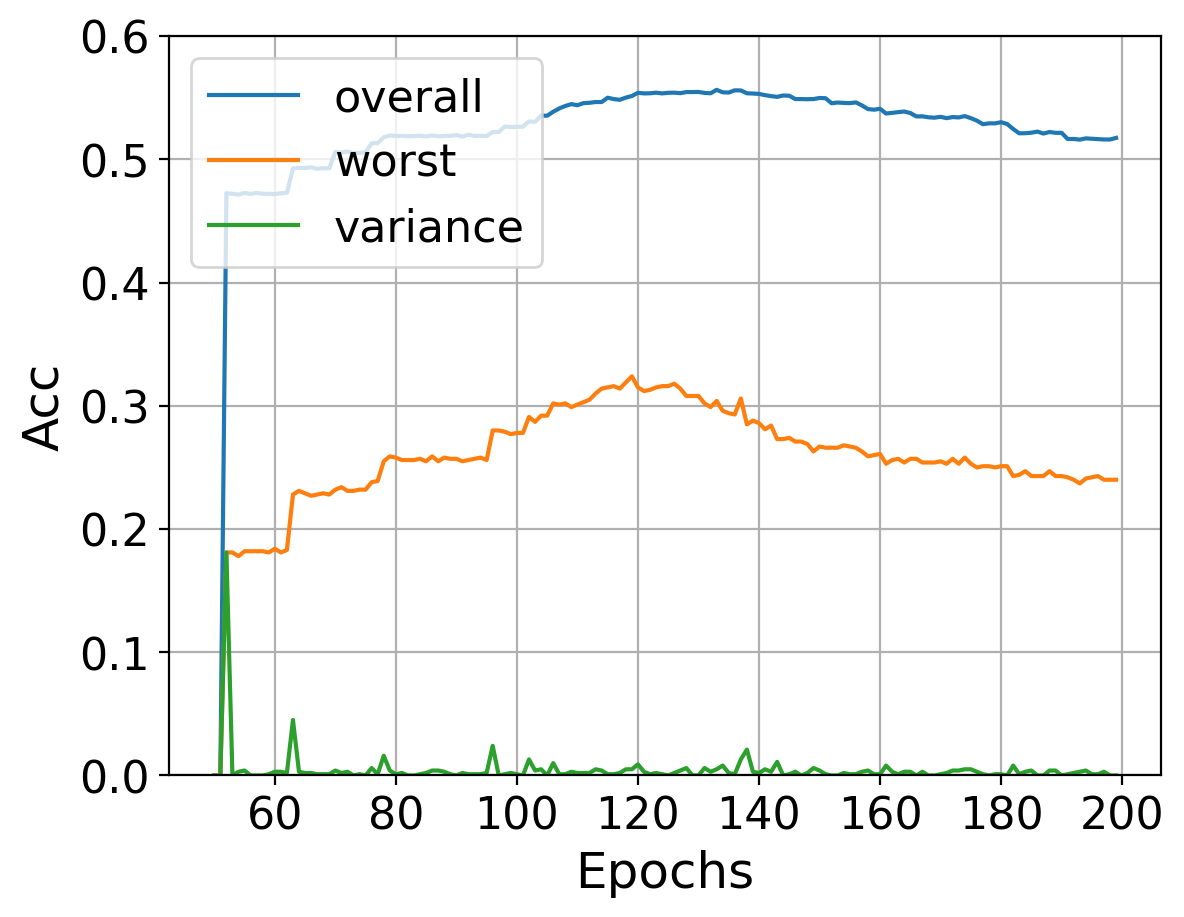}
        \\
        (a) & (b)
    \end{tabular}
    \vspace{-0.15 in}
    \caption{Comparison of overall robustness, the worst class robustness, and the absolute variation of the worst class robustness between adjacent checkpoints. (a): Vanilla AT. (b): AT with fairness aware weight averaging (FAWA), start from epoch 50.}
    \label{fig:worst and overall}
\end{figure}

This result is consistent with the Theorem \ref{theorem:compare}. Recall that in the toy model, hard class $y=-1$ costs more clean accuracy to exchanges for little robustness improvement than easy class $y=+1$. Therefore, similar to the analysis on perturbation margin $\epsilon$, we also point out that there exists a proper $\beta_y$ for each class.

\subsection{Fluctuation Effect}
\label{fluctuate}
In this section, we reveal an intriguing property regarding the fluctuation of class-wise robustness during adversarial training. In Fig.~\ref{fig:worst and overall}(a), we plot the overall robustness, the worst class robustness, and the variance of the worst robustness between adjacent epochs in vanilla adversarial training. While the overall robustness tends to be more stable between adjacent checkpoints (except when the learning rate decays), the worst class robustness fluctuates significantly. Particularly, many adjacent checkpoints between the $101-120$-th epochs exhibit a nearly 10\% difference in the worst class robustness, while changes in overall robustness are negligible (less than 1\%).
Therefore, previously widely used selecting the best checkpoint based on overall robustness may result in an extremely unfair model. Taking the plotted training process as an example, the model achieves the highest robust accuracy of 53.2\% at the 108-th epoch, which only has 23.5\% robust accuracy on the worst class. In contrast, the checkpoint at epoch 110, which has 52.6\% overall and 28.1\% worst class robust accuracy, is preferred when considering fairness.

\section{Class-wise Calibrated Fair Adversarial Training}
With the above analysis, we introduce our proposed \textbf{C}lass-wise  calibrated \textbf{F}air \textbf{A}dversarial training (CFA) framework in this section.
Overall, the CFA framework consists of three main components: Customized Class-wise perturbation Margin (CCM), Customized Class-wise Regularization (CCR), and Fairness Aware Weight Averaging (FAWA).
The CCM and CCR customize appropriate training configurations for different classes, and FAWA modifies weight averaging to improve and stabilize fairness.

\subsection{Class-wise Calibrated Margin (CCM)}
\label{CCM}
In Sec.~\ref{margin analysis}, we have demonstrated that different classes prefer specific perturbation margin $\epsilon$ in adversarial training. However, it is impractical to directly find the optimal class-wise margin. Inspired by a series of instance-wise adaptive adversarial training approaches~\cite{ding2018mma,wang2019improving,balaji2019instance}, which customize train setting for each instance according to the model performance on current example, we propose to leverage the class-wise training accuracy as the measurement of difficulty.

Suppose the $k$-th class achieved train robust accuracy $t_k\in[0,1]$ in the last training epoch. 
In the next epoch, we aim to update the margin $\epsilon_k$ for class $k$ based on $t_k$.
Based on our analysis in Sec.~\ref{margin analysis}, we consider using a relatively smaller margin for the hard classes which are more vulnerable to attacks, and identify the \textit{difficulty} among classes by the train robust accuracy tracked from the previous epoch.
To avoid $\epsilon_k$ too small, we add a hyper-parameter $\lambda_1$ (called \textit{base perturbation budget}) on all $t_k$ and set the calibrated margin $\epsilon_k$ by multiply the coefficient on primal margin $\epsilon$:
\begin{equation}
    \epsilon_k\gets (\lambda_1 +  t_k)\cdot \epsilon,
    \label{ccm formula}
\end{equation}
where $\epsilon$ is the original perturbation margin, \textit{e.g.}, $8/255$ that is commonly used for CIFAR-10 dataset. 
Note that the calibrated margin $\epsilon_k$ can adaptively converge to find the proper range during the training phase, for example, 
 if the margin is too small for class $k$, the model will perform high train robust accuracy $t_k$ and then increase $\epsilon_k$ by schedule (\ref{ccm formula}). 

\subsection{Class-wise Calibrated Regularization (CCR)}
We further customize different robustness regularization $\beta$ of TRADES for different classes.
Recall the objective function (\ref{TRADES}) of TRADES, we hope the hard classes tend to bias more weight on its clean accuracy. Still, we measure the difficulty by the train robust accuracy $t_k$ for class $k$, 
and propose the following calibrated robustness regularization $\beta_k$:
\begin{equation}
    \beta_k\gets (\lambda_2 + t_k) \cdot \beta.
\end{equation}
where $\beta$ is the originally selected parameter.
The objective function (\ref{TRADES}) can be rewritten as:
\begin{equation}
\label{ccw objective}
    \mathcal L_{\boldsymbol\theta}(\beta;x,y)=
    \frac{\mathcal L(\boldsymbol\theta;x,y) + \beta_y\max\limits_{\|{x'}-{x}\|\le\epsilon} \mathcal{K} (f_{\boldsymbol\theta}({x}), f_{\boldsymbol\theta}({x'}))}{1+\beta_y}.
\end{equation}

To balance the weight between different classes, we add a denominator $1+\beta_y$ since $\beta_y$ is distinct among classes. Therefore, for the hard classes which have lower $\beta_y$ tend to bias higher weight $\frac 1 {1+\beta_y}$ on its natural loss $\mathcal L(\boldsymbol\theta;x,y)$. Note that simply replacing $\epsilon$ in (\ref{ccw objective}) with $\epsilon_k$ can combine the calibrated margin with this calibrated regularization.  On the other hand, for general adversarial training algorithms, our calibrated margin schedule (\ref{ccm formula}) can also be combined.

\subsection{Fairness Aware Weight Average (FAWA)}
\label{sec: fawa}
As plotted in Fig.~\ref{fig:worst and overall}(a), the worst class robustness changes largely, among which part of checkpoints performs extremely poor robust fairness. 
Previously, there are a series of weight averaging methods to make the model training stable, \textit{e.g.}, exponential moving average (EMA) \cite{DBLP:conf/uai/IzmailovPGVW18,wang2022self}, thus we hope to further improve the worst class robustness by fixing the weight average algorithm.

Inspired by the large fluctuation of the robustness fairness among checkpoints, we consider eliminating the \textit{unfair} checkpoints out in the weight averaging process. To this end, we propose a \textit{Fairness Aware Weight Average (FAWA)} approach, which sets a threshold $\delta$ on the worst class robustness of the new checkpoint in the EMA process. Specifically, we extract a validation set from the dataset, and each checkpoint is adopted in the weight average process if and only if its worst class robustness is higher than $\delta$. Fig.~\ref{fig:worst and overall}(b) shows the effect of the proposed FAWA. The difference between adjacent epochs is extremely small (less than 1\%), and the overall robustness also outperforms vanilla AT.

\begin{algorithm}[h]
    \caption{TRADES with CFA}
	\label{alg:cfa}
	\KwIn{A DNN classifier $f_{\boldsymbol\theta}(\cdot)$ with parameter $\boldsymbol\theta$; Train dataset $D=\{(x_i, y_i)\}_{i=1}^N$; Batch size $m$; Initial perturbation margin $\epsilon$ and robustness regularization $\beta$; Train epochs $N$; Batch size $m$; Learning rate $\eta$; Weight average decay rate $\alpha$; Fairness threshold $\delta$}
	\KwOut{A fair and robust DNN classifier $\bar{f}_{\bar{\boldsymbol\theta}}(\cdot)$}
	\tcc{Initialize parameters and datasets}
    Initialize $\boldsymbol\theta\gets\boldsymbol\theta_0, \bar{\boldsymbol\theta}\gets\boldsymbol\theta$\;
    Split $D=D_{\text{train}}\cup D_{\text{valid}}$\;
    \For{$y\in\mathcal Y$}{
    \tcc{Initialize $\epsilon_y$ and $\beta_y$}
    $\epsilon_y\gets\epsilon, \beta_y\gets\beta$\;
    }
    
    \For{$T\gets 1,2,\cdots N$}{
        \For{Every minibatch $(x,y)$ in $D_{\text{train}}$}{
        \tcc{Use $\epsilon_y$ and $\beta_y$ to train}
        $x'\gets \arg\max\limits_{x'\in\mathcal{B}(x,\epsilon_y)}\mathcal K (f_{\boldsymbol\theta}({x}), f_{\boldsymbol\theta}({x'}))$\;
        $\boldsymbol\theta \gets \boldsymbol\theta - \eta\nabla_{\boldsymbol\theta}\mathcal L_{\boldsymbol\theta}(\beta_y;x,y)$\;
        }
        \For{$y\in\mathcal Y$}{
        $t_y\gets Train\_Acc(f_\theta, T)$\;
        \tcc{Update $\epsilon_y,\beta_y$ with $t_y$}
        $\epsilon_y\gets (\lambda_1 + t_k)\cdot\epsilon$\;
        $\beta_y\gets (\lambda_2 + t_k)\cdot \epsilon$\;}
        \tcc{Fairness Aware Weight Average}
        \If{$\min_{y\in\mathcal Y} \mathcal{R}_{y}(f_{\boldsymbol\theta}, D_{\text{valid}})\ge \delta$}{
        \vspace{0.1cm}
        $\bar{\boldsymbol{\theta}}\gets \alpha\bar{\boldsymbol\theta}+(1-\alpha){\boldsymbol\theta}$\;}
    }
    \textbf{return} $\bar{f}_{\bar{\boldsymbol\theta}}$\;

\end{algorithm}

\subsection{Discussion}

Overall, by combining the above components, we accomplish our CFA framework. An illustration of incorporating CFA to TRADES is shown in Alg.~\ref{alg:cfa}. Note that for other methods like AT, we can still incorporate CFA by removing the CCR schedule specified for TRADES. Moreover, we discuss the difference between our proposed CFA and other works. 

\noindent \textbf{Comparison with Fair Robust Learning (FRL)~\cite{xu2021robust}.}
Here we highlight the differences between our CFA framework and Fair Robust Learning (FRL), the only existing adversarial training algorithm designed to improve the fairness of class-wise robustness. The FRL framework consists of two components: remargin and reweight. Initially, a robust model is trained, and a fairness constraint on the difference of robustness among classes is set. When the constraint is violated, the model is fine-tuned persistently by increasing the perturbation bound $\epsilon_k$ and weighting the loss of the hard classes. Although CFA also includes adaptive margin and regularization weight schedules, our work is fundamentally distinct from FRL. Firstly, as discussed in Sec.~\ref{margin analysis}, a larger margin only mitigates the robust over-fitting problem but does not provide higher peak performance. In contrast, our approach aims to customize the proper margin for each class, which boosts the best performance. Secondly, FRL improves robust fairness at the cost of reducing overall robustness, which could be seen as \textit{unfair} to other classes. However, our CFA framework improves both overall and worst class performance. In addition, FRL requires an initial robust model before fairness fine-tuning, resulting in extra computational burden. Finally, the fluctuation effect discussed in Sec.~\ref{fluctuate} is not considered in FRL. 

\noindent \textbf{Comparison with Instance-wise Adversarial Training.}
\label{dis:instance}
Though there exists a series of instance-wise adaptive adversarial training~\cite{ding2018mma,  balaji2019instance,wang2019improving,cheng2020cat, zhang2020attacks, GAIR, cai2018curriculum, wang2021convergence} toward better robust generalization, to the best of our knowledge, we are the first work to pursue this from a class-wise perspective. 
Here we demonstrate several differences between our class-wise and other instance-wise adversarial training algorithms. 
First of all, CFA focuses on improve both overall and the worst class robust accuracy, while all existing instance-wise approaches only focus on overall robustness. Unfortunately, as shown in Sec.~\ref{sec:experiment}, the instance-wise ones are not comparable with our CFA from the perspective of fairness. 
In addition, instance-wise methods can be seen as to find the solution for each individual sample, while class-wise ones are to find the solution for multiple samples. Thus, class-wise methods can alleviate the frequent fluctuation while remaining the specificity (a class of samples) of configurations among training samples. Therefore, our class-wise calibration achieves a better trade-off between flexibility and stability. Finally,  some instance-wise approaches can be well-combined with our CFA framework to further boost their performance.

\section{Experiment}
\label{sec:experiment}
In this section, we demonstrate the effectiveness of our proposed CFA framework to improve both overall and class-wise robustness. 

\begin{table*}[!t]
    \centering
    \small
        \caption{Overall comparison of our proposed CFA framework with original methods.}
    \begin{tabular}{l|cc|cc}
    \toprule[2pt]
        & \multicolumn{2}{c|}{Best (Avg. / Worst)} & \multicolumn{2}{c}{Last (Avg. / Worst)} 
        \\
        \textbf{Method} & Clean Accuracy  & AA. Accuracy &
        Clean Accuracy  & AA. Accuracy 
        \\ \midrule[1pt]
        AT & 
        \textbf{82.3} {\scriptsize  $\pm 0.8$ } 
        /
        63.9 {\scriptsize  $\pm1.6$ }
        &  
        46.7 {\scriptsize  $\pm0.5$ }
        /
        20.1 {\scriptsize $\pm1.3$}
        &
        84.1 {\scriptsize  $\pm0.2$ }
        /
        65.1{\scriptsize  $\pm 2.4$ }
        &
        43.0 {\scriptsize  $\pm0.4$ }
        /
        15.5 {\scriptsize  $\pm1.8$ }
         \\
         AT + EMA &
        81.9 {\scriptsize  $\pm 0.3$ }
         /
        61.6 {\scriptsize  $\pm0.5$ }
        &
        49.6 {\scriptsize  $\pm0.2$ }
        /
        21.3 {\scriptsize  $\pm0.8$ }
        &
        \textbf{84.8} {\scriptsize  $\pm0.1$ }
        /
        67.7 {\scriptsize  $\pm0.7$ }
        &
        44.3 {\scriptsize  $\pm0.5$ }        
        /
        18.1 {\scriptsize  $\pm0.5$ }
        \\
        \textbf{AT + CFA} &
        80.8 {\scriptsize  $\pm0.3$ }
        /
        \textbf{64.6} {\scriptsize  $\pm0.4$ }
        &
        \textbf{50.1} {\scriptsize  $\pm0.3$ }
        /
        \textbf{24.4} {\scriptsize  $\pm0.3$ }
        &
        83.6 {\scriptsize  $\pm0.2$ }
        /
        \textbf{68.7} {\scriptsize  $\pm0.7$ }
        &
        \textbf{47.7} {\scriptsize  $\pm0.4$ }
        /
        \textbf{20.5} {\scriptsize  $\pm0.4$ }
        \\ \midrule[1pt]
        TRADES &
        \textbf{82.3} {\scriptsize  $\pm0.1$ }
        /
        \textbf{67.8} {\scriptsize  $\pm0.6$ }
        &
        48.3 {\scriptsize  $\pm0.3$ }
        /
        21.7 {\scriptsize  $\pm0.5$ }
        &
        83.9 {\scriptsize  $\pm0.3$ }
        /
        66.9 {\scriptsize  $\pm1.5$ }
        &
        46.9 {\scriptsize  $\pm0.3$ }
        /
        18.5 {\scriptsize  $\pm1.3$ }
        \\
        TRADES + EMA & 
        81.2 {\scriptsize  $\pm0.4$ }
        /
        65.0 {\scriptsize  $\pm0.7$ }
        &
        49.7 {\scriptsize  $\pm0.3$ }
        /
        24.2 {\scriptsize  $\pm0.6$ }
        &
        \textbf{84.5} {\scriptsize  $\pm0.1$ }
        /
        67.9 {\scriptsize  $\pm0.1$ }
        &
        48.3 {\scriptsize  $\pm0.2$ }
        /
        20.7 {\scriptsize  $\pm0.3$ }
        \\
        \textbf{TRADES + CFA} &
        80.4 {\scriptsize  $\pm0.2$ }
        /
        66.2 {\scriptsize  $\pm0.5$ }
        &
        \textbf{50.1} {\scriptsize  $\pm0.2$ }
        /
        \textbf{26.5} {\scriptsize  $\pm0.4$ }
        &
        83.0 {\scriptsize  $\pm0.1$ }
        /
        \textbf{68.1} {\scriptsize  $\pm0.3$ }
        &
        \textbf{49.3} {\scriptsize  $\pm0.1$ }
        /
        \textbf{21.5} {\scriptsize  $\pm0.3$ }
        \\ \midrule[1pt]

        FAT &
        84.6 {\scriptsize  $\pm0.4$ }
        /
        \textbf{69.2} {\scriptsize  $\pm 0.8$ }
        &
        45.7 {\scriptsize  $\pm0.6$ }
        /
        17.2 {\scriptsize  $\pm1.3$ }
        &
        85.4 {\scriptsize  $\pm0.2$ }
        /
        70.8 {\scriptsize  $\pm1.9$ }
        &
        42.1 {\scriptsize  $\pm0.1$ }
        /
        14.8 {\scriptsize  $\pm1.6$ }
        \\
        FAT + EMA & 
        \textbf{85.2} {\scriptsize  $\pm0.2$ }
        /
        66.7 {\scriptsize  $\pm0.6$ }
        &
        48.6 {\scriptsize  $\pm0.1$ }
        /
        18.3 {\scriptsize  $\pm0.5$ }
        &
        \textbf{85.7} {\scriptsize  $\pm0.2$ }
        /
        \textbf{71.2} {\scriptsize  $\pm0.4$ }
        &
        43.2 {\scriptsize  $\pm0.1$ }
        /
        15.7 {\scriptsize  $\pm0.7$ }
        \\
        \textbf{FAT + CFA} & 
        82.1 {\scriptsize  $\pm0.3$ }
        /
        64.7 {\scriptsize  $\pm0.9$ }
        &
        \textbf{49.6} {\scriptsize  $\pm 0.1$ }
        /
        \textbf{20.9} {\scriptsize  $\pm0.8$ }
       &
       84.3 {\scriptsize  $\pm0.1$ }
       /
       69.4 {\scriptsize  $\pm0.3$ }
       &
       \textbf{45.1} {\scriptsize  $\pm0.2$ }
       /
       \textbf{16.7} {\scriptsize  $\pm0.2$ }
        \\ \midrule[1pt]
        FRL & 
        {82.8} {\scriptsize $\pm  0.1 $}
        /
        \textbf{71.4} {\scriptsize $\pm  2.4 $}
        &
        45.9 {\scriptsize $\pm  0.3 $}
        /
        25.4 {\scriptsize $\pm  2.0 $}
        &
        \textbf{82.8} {\scriptsize $\pm  0.2 $}
        /
        72.9 {\scriptsize $\pm  1.5 $}
        &
        44.7 {\scriptsize $\pm  0.2 $}
        /
        23.1 {\scriptsize $\pm  0.8 $}
        \\
        FRL + EMA &
        \textbf{83.6} {\scriptsize  $\pm0.3$ }
        /
        69.5 {\scriptsize  $\pm0.7$ }
        &
        \textbf{46.1} {\scriptsize  $\pm0.2$ }
        /
        \textbf{25.6} {\scriptsize  $\pm0.4$ }
        &
        {81.9} {\scriptsize  $\pm0.2$ }
        /
        \textbf{74.2} {\scriptsize  $\pm0.3$ }
        &
        \textbf{44.9} {\scriptsize  $\pm0.2$ }
        / 
        \textbf{24.5} {\scriptsize  $\pm0.3$ }
        \\
        
     \bottomrule[2pt]
    \end{tabular}

    \label{tab:overall}

\end{table*}

\subsection{Experimental Setup}
We conduct our experiments on the benchmark dataset CIFAR-10~\cite{krizhevsky2009learning} using PreActResNet-18 (PRN-18)~\cite{he2016identity} model. Experiments on Tiny-ImageNet can be found in Appendix~\ref{C1}.

\noindent \textbf{Baselines.} We select vanilla adversarial training (AT)~\cite{madry2017towards} and TRADES~\cite{zhang2019theoretically} as our baselines.
Additionally, since our {Fairness Aware Weight Average (FAWA)} method is a variant of the weight average method with \textit{Exponential Moving Average (EMA)}, we include baselines with EMA as well.
For instance-wise adaptive adversarial training approaches, we include FAT~\cite{zhang2020attacks}, which adaptively adjusts attack strength on each instance. Finally, we compare our approach with FRL~\cite{xu2021robust}, the only existing adversarial training algorithm that focuses on improving the fairness of class-wise robustness.

\noindent \textbf{Training Settings.} Following the best settings in ~\cite{rice2020overfitting}, we train a PRN-18 using SGD with momentum 0.9, weight decay $5\times 10^{-4}$, and initial learning rate 0.1 for 200 epochs. The learning rate is divided by $10$ after epoch 100 and 150. All experiments are conducted by default perturbation margin $\epsilon=8/255$, and for TRADES, we initialize $\beta=6$. For the base attack strength for Class-wise Calibrated Margin (CCM), we set $\lambda_1=0.5$ for AT and $\lambda_1=0.3$ for TRADES since the training robust accuracy of TRADES is higher than AT. For FAT, we set $\lambda_1=0.7$ to avoid the attack being too weak to hard classes. Besides, we set $\lambda_2=0.5$ for Class-wise Calibrated Regularization (CCR) in TRADES. For the weight average methods, the decay rate of FAWA and EMA is set to 0.85, and the weight average processes begin at the 50-th epoch for better initialization. We draw 2\% samples from each class as the validation set for FAWA, and train on the rest of 98\% samples, hence FAWA does not lead to extra computational costs.
The fairness threshold for FAWA is set to 0.2. 

\noindent \textbf{Metrics.} We evaluate the clean and robust accuracy both in average and the worst case among classes. The robustness is evaluated by \textbf{AutoAttack (AA)}~\cite{croce2020reliable}, a well-known reliable attack for robustness evaluation.
To perform the best performance during the training phase, we adopt early stopping in adversarial training~\cite{rice2020overfitting} and present both the best and last results among training checkpoints. Further, as discussed in Sec.~\ref{fluctuate} that the worst class robust accuracy changes drastically, we select the checkpoint that achieves the highest sum of overall and the worst class robustness to report the results for a fair comparison.

\subsection{Robustness and Fairness Performance}
We implement our proposed training configuration schedule on AT, TRADES, and FAT. To evaluate the effectiveness of our approach, we conduct five independent experiments for each method and report the mean result and standard deviation.

As summarized in Table~\ref{tab:overall}, CFA helps each method achieve a significant robustness improvement both in average and the worst class at the best and last checkpoints.
Furthermore, when compared with baselines that use weight average (EMA), our CFA still achieves higher overall and the worst class robustness for each method, especially in the worst class at the best checkpoints, where the improvement exceeds 2\%. 
Note that the vanilla FAT only achieves 17.2\% the worst class robustness at the best checkpoint which is even lower than TRADES, which verifies the discussion in Sec. \ref{dis:instance} that instance-wise adaptive approaches are not helpful for robustness fairness. We also visualize and compare the robustness for each class in Appendix~\ref{C2}, which shows that CFA indeed reduces the difference among class-wise robustness and improves the fairness without harming other classes. 

We also compare our approach with FRL~\cite{xu2021robust}. However, since FRL also applies a remargin schedule, we cannot incorporate our CFA into FRL.
Therefore, we only report results of FRL with and without EMA in Table~\ref{tab:overall}.
As FRL is a variant of TRADES that applies the loss function of TRADES, we compare the results of FRL with TRADES and TRADES+CFA.
From Table~\ref{tab:overall}, we observe that FRL and FRL+EMA show only marginal progress (less than 2\%) in the worst class robustness as compared to TRADES+EMA, but at a expensive cost (about 3\%) of reducing the average performance.
As demonstrated in Sec.~\ref{margin analysis}, larger margin which is adopted in FRL mainly mitigates the robust over-fitting issue but does not bring satisfactory best performance. 
This is further confirmed by the performance of final checkpoints of FRL,
where FRL exhibits better performance in the worst class robustness.
In contrast, we calibrate the appropriate margin for each class rather than simply enlarging them, thus achieving both better robustness and fairness at the best checkpoint, \textit{i.e.}, our TRADES+CFA outperforms FRL+EMA in both average (about 4\%) and the worst class (about 1\%) robustness.

\subsection{Ablation Study}
In this section, we show the usefulness of each component of our CFA framework. Note that we still apply \textbf{AutoAttack (AA)} to evaluate robustness.

\subsubsection{Effectiveness of Calibrated Configuration}
\label{ablation}
First, we compare our calibrated adversarial configuration including CCM $\epsilon_y$ and CCR $\beta_y$ with vanilla ones for AT, TRADES, and FAT. As Table~\ref{tab:calibrated conf} shows, both the average and worst class robust accuracy are improved for all three methods by applying CCM. Besides, CCR, which is customized for TRADES, also improves the performance of vanilla TRADES. All experiments verify that our proposed class-wise adaptive adversarial configurations are effective for robustness and fairness improvement.

\begin{figure}[!t]
    \centering
    \begin{tabular}{cc}
        \includegraphics[width=0.22\textwidth]{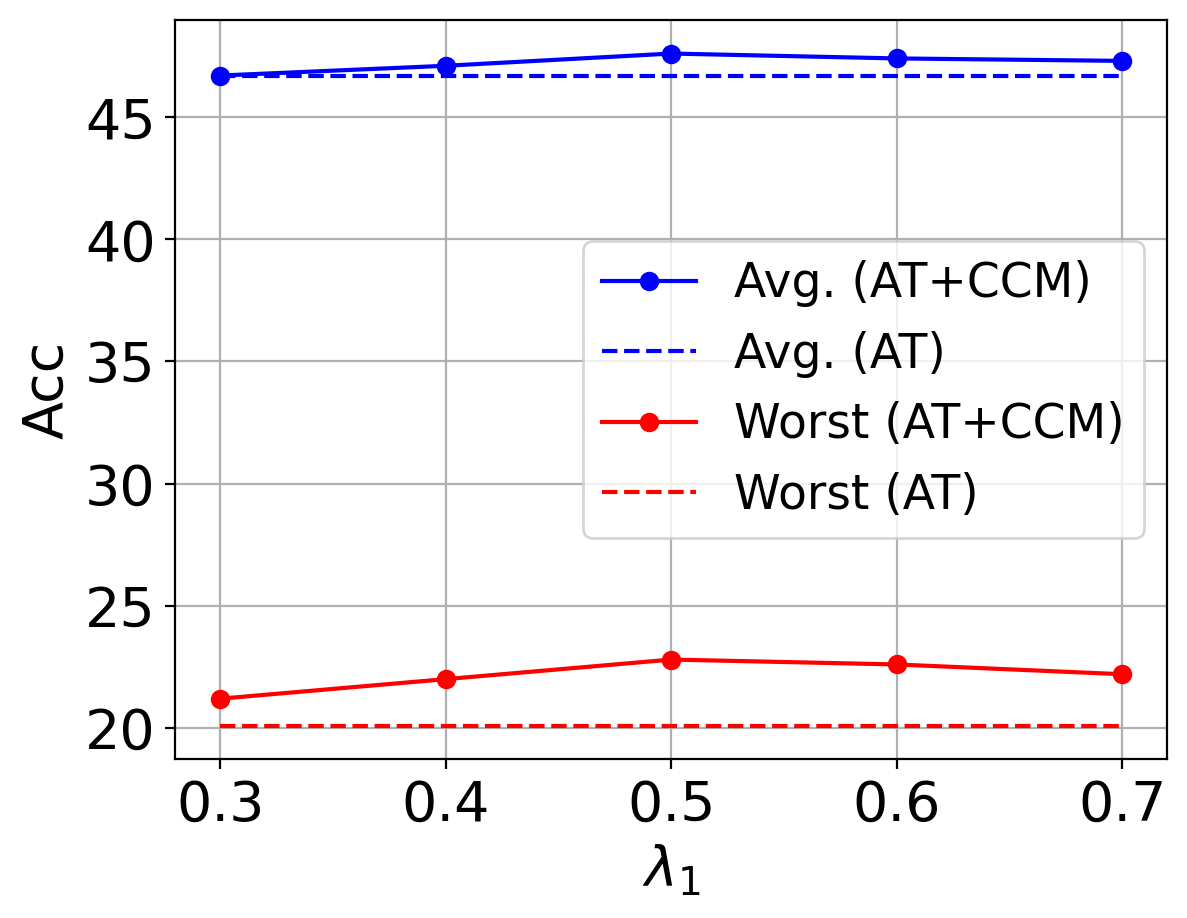}  &  \includegraphics[width=0.22\textwidth]{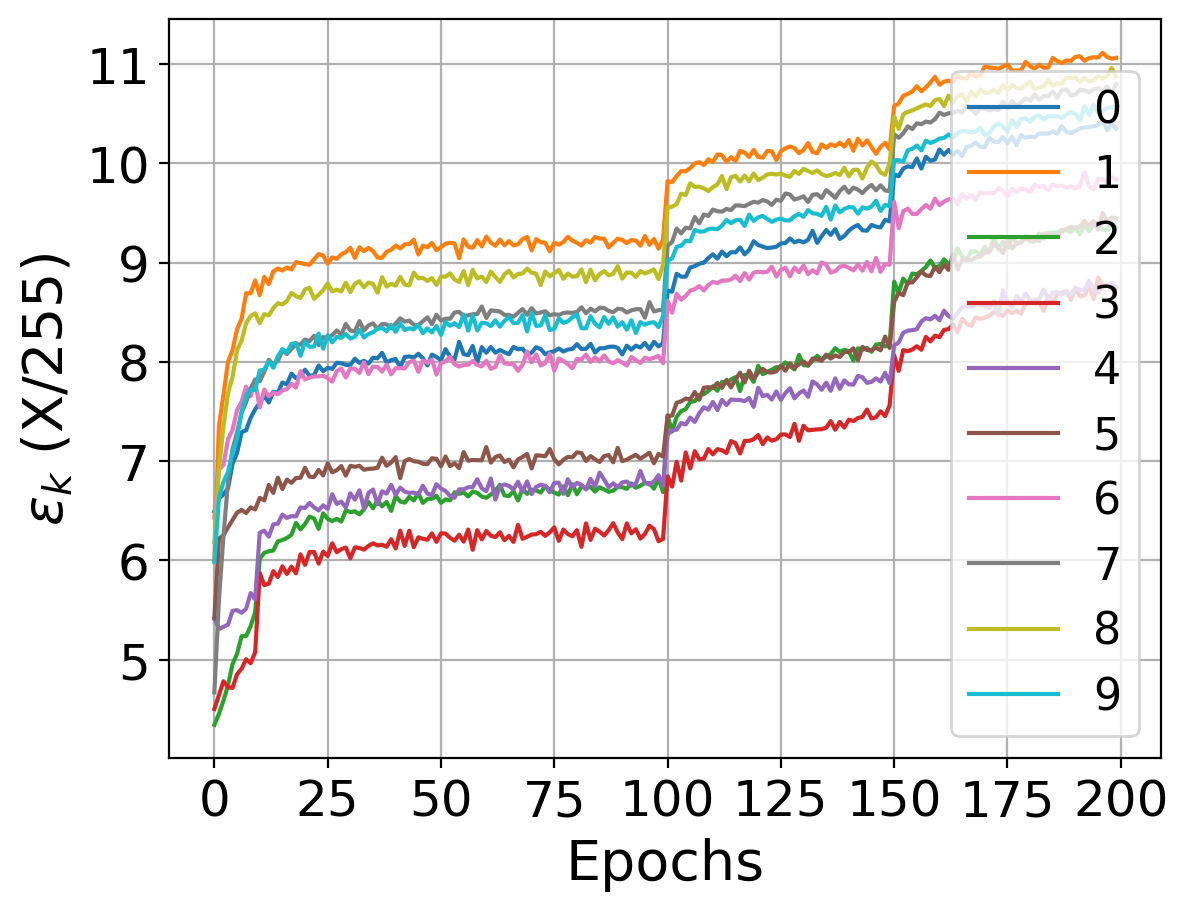}
         \\
        (a) & (b)
    \end{tabular}
    \caption{Analysis on the base perturbation budget $\lambda_1$. (a): Average and the worst class robustness of models trained with different $\lambda_1$ (solid) and vanilla AT (dotted). (b): Class-wise calibrated margin $\epsilon_k$ in the training phase of $\lambda_1=0.5$.}
    \label{fig:ccm analysis}
\end{figure}

We also investigate the influence of base perturbation budget $\lambda_1$ by conducting 5 experiments of AT incorporated CCM with $\lambda_1$ varies from 0.3 to 0.7. The comparison is plotted in Fig.~\ref{fig:ccm analysis}(a). We can see that all models with different $\lambda_1$ show better overall and the worst class robustness than vanilla AT, among which $\lambda_1=0.5$ performs best. 
We can say that CCM has satisfactory adaptive ability on adjusting $\epsilon_k$ and is not heavily rely on the selection of $\lambda_1$. Fig.~\ref{fig:ccm analysis}(b) shows the class-wise margin used in the training phase for $\lambda_1=0.5$. We can see the hard classes (class 2,3,4,5) use smaller $\epsilon_k$ than the original $\epsilon=8/255$, while the easy classes use larger ones, which is consistent with our empirical observation on different margins in Sec.~\ref{margin analysis} and can explain why CCM is helpful to improve performance. We also present a similar comparison experiments on $\lambda_2$ for CCR in Appendix~\ref{C3}.

\subsubsection{FAWA Improves Worst Class Robustness}
Here we present the results of our Fairness Aware Weight Averaging (FAWA) compared with the simple EMA method in Table~\ref{tab:fawa}. 
By eliminating the unfair checkpoints out,  our FAWA achieves significantly better performance than EMA on the worst class robustness (nearly 2\% improvement) with negligible decrease on the overall robustness (less than 0.3\%). This verifies the effectiveness of FAWA on improving robustness fairness.

\begin{table}[!t]
    \centering
        \caption{Comparison of models with/without our class-wise calibrated configurations including margin $\epsilon$ and regularization $\beta$.}
    \begin{tabular}{l|cc}
        \toprule[1.5pt]
        \textbf{Method} & Avg. Robust & Worst Robust \\ 
        \midrule[1pt]
        AT  &  46.7 & 20.1\\
        + CCM &  \textbf{47.6} & \textbf{22.8}\\ \midrule
        TRADES & 48.3 & 21.7\\
        + CCM & 48.4 & 22.5\\
        + CCR & {48.9} & 23.5\\
        + CCM + CCR &  \textbf{49.2} & \textbf{23.8} \\ \midrule
        FAT & 45.7 & 17.2 \\
        + CCM & \textbf{46.8} & \textbf{18.9} \\
        \bottomrule[1.5pt]
    \end{tabular}

    \label{tab:calibrated conf}
\end{table}

\begin{table}[!t]
    \centering
        \caption{Comparison of simple EMA and our FAWA.}
    \begin{tabular}{l|cc}
    \toprule[1.5pt]
     Method & Avg. Robust & Worst Robust
     \\ \midrule[1pt]
        AT + EMA &  \textbf{49.6} & 21.3\\
        AT + FAWA &  49.3 & \textbf{23.1} \\ \midrule
        TRADES + EMA & \textbf{49.7} & 24.2 \\
        TRADES + FAWA & 49.4 & \textbf{25.1} \\ \midrule
        FAT + EMA & \textbf{48.6} & 18.3\\
        FAT + FAWA & 48.5 & \textbf{19.9}\\
    \bottomrule[1.5pt]
    \end{tabular}

    \label{tab:fawa}
\end{table}

\section{Conclusion}
In this paper, we first give a theoretical analysis of how attack strength in adversarial training impacts the performance of different classes. Then, we empirically show the influence of adversarial configurations on class-wise robustness and the fluctuate effect of robustness fairness, and point out there should be some appropriate configurations for each class.
Based on these insights, we propose a \textbf{C}lass-wise calibrated \textbf{F}air \textbf{A}dversarial training (CFA) framework to adaptively customize class-wise train  configurations for improving robustness and fairness.
Experiment shows our CFA outperforms state-of-the-art methods both in overall and fairness metrics, and can be easily incorporated into existing methods to further enhance their performance. 

\section*{Acknowledgement}
Yisen Wang is partially supported by the National Key R\&D Program of China (2022ZD0160304), the National Natural Science Foundation of China (62006153), and Open Research Projects of Zhejiang Lab (No. 2022RC0AB05).

{\small
\bibliographystyle{ieee_fullname}
\bibliography{egbib}

\begin{thebibliography}{10}\itemsep=-1pt

\bibitem{athalye2018obfuscated}
Anish Athalye, Nicholas Carlini, and David Wagner.
\newblock Obfuscated gradients give a false sense of security: Circumventing
  defenses to adversarial examples.
\newblock In {\em ICML}, 2018.

\bibitem{bai2019hilbert}
Yang Bai, Yan Feng, Yisen Wang, Tao Dai, Shu-Tao Xia, and Yong Jiang.
\newblock Hilbert-based generative defense for adversarial examples.
\newblock In {\em ICCV}, 2019.

\bibitem{balaji2019instance}
Yogesh Balaji, Tom Goldstein, and Judy Hoffman.
\newblock Instance adaptive adversarial training: Improved accuracy tradeoffs
  in neural nets.
\newblock {\em arXiv preprint arXiv:1910.08051}, 2019.

\bibitem{benz2021robustness}
Philipp Benz, Chaoning Zhang, Adil Karjauv, and In~So Kweon.
\newblock Robustness may be at odds with fairness: An empirical study on
  class-wise accuracy.
\newblock {\em arXiv preprint arXiv:2010.13365}, 2020.

\bibitem{cai2018curriculum}
Qi-Zhi Cai, Min Du, Chang Liu, and Dawn Song.
\newblock Curriculum adversarial training.
\newblock {\em arXiv preprint arXiv:1805.04807}, 2018.

\bibitem{chen2015deepdriving}
Chenyi Chen, Ari Seff, Alain Kornhauser, and Jianxiong Xiao.
\newblock Deepdriving: Learning affordance for direct perception in autonomous
  driving.
\newblock In {\em CVPR}, 2015.

\bibitem{cheng2020cat}
Minhao Cheng, Qi Lei, Pin-Yu Chen, Inderjit Dhillon, and Cho-Jui Hsieh.
\newblock Cat: Customized adversarial training for improved robustness.
\newblock {\em arXiv preprint arXiv:2002.06789}, 2020.

\bibitem{croce2020reliable}
Francesco Croce and Matthias Hein.
\newblock Reliable evaluation of adversarial robustness with an ensemble of
  diverse parameter-free attacks.
\newblock In {\em ICML}, 2020.

\bibitem{das2017keeping}
Nilaksh Das, Madhuri Shanbhogue, Shang-Tse Chen, Fred Hohman, Li Chen,
  Michael~E Kounavis, and Duen~Horng Chau.
\newblock Keeping the bad guys out: Protecting and vaccinating deep learning
  with jpeg compression.
\newblock {\em arXiv preprint arXiv:1705.02900}, 2017.

\bibitem{ding2018mma}
Gavin~Weiguang Ding, Yash Sharma, Kry Yik~Chau Lui, and Ruitong Huang.
\newblock Mma training: Direct input space margin maximization through
  adversarial training.
\newblock {\em arXiv preprint arXiv:1812.02637}, 2018.

\bibitem{goodfellow2014explaining}
Ian~J Goodfellow, Jonathon Shlens, and Christian Szegedy.
\newblock Explaining and harnessing adversarial examples.
\newblock {\em arXiv preprint arXiv:1412.6572}, 2014.

\bibitem{he2016identity}
Kaiming He, Xiangyu Zhang, Shaoqing Ren, and Jian Sun.
\newblock Identity mappings in deep residual networks.
\newblock In {\em ECCV}, 2016.

\bibitem{DBLP:conf/uai/IzmailovPGVW18}
Pavel Izmailov, Dmitrii Podoprikhin, Timur Garipov, Dmitry~P. Vetrov, and
  Andrew~Gordon Wilson.
\newblock Averaging weights leads to wider optima and better generalization.
\newblock In {\em UAI}, 2018.

\bibitem{krizhevsky2009learning}
Alex Krizhevsky, Geoffrey Hinton, et~al.
\newblock Learning multiple layers of features from tiny images.
\newblock 2009.

\bibitem{ma2019understanding}
Xingjun Ma, Yuhao Niu, Lin Gu, Yisen Wang, Yitian Zhao, James Bailey, and Feng
  Lu.
\newblock Understanding adversarial attacks on deep learning based medical
  image analysis systems.
\newblock {\em Pattern Recognition}, 2020.

\bibitem{madry2017towards}
Aleksander Madry, Aleksandar Makelov, Ludwig Schmidt, Dimitris Tsipras, and
  Adrian Vladu.
\newblock Towards deep learning models resistant to adversarial attacks.
\newblock {\em arXiv preprint arXiv:1706.06083}, 2017.

\bibitem{mo2022adversarial}
Yichuan Mo, Dongxian Wu, Yifei Wang, Yiwen Guo, and Yisen Wang.
\newblock When adversarial training meets vision transformers: Recipes from
  training to architecture.
\newblock In {\em NeurIPS}, 2022.

\bibitem{papernot2016distillation}
Nicolas Papernot, Patrick McDaniel, Xi Wu, Somesh Jha, and Ananthram Swami.
\newblock Distillation as a defense to adversarial perturbations against deep
  neural networks.
\newblock In {\em SP}, 2016.

\bibitem{rice2020overfitting}
Leslie Rice, Eric Wong, and Zico Kolter.
\newblock Overfitting in adversarially robust deep learning.
\newblock In {\em ICML}, 2020.

\bibitem{szegedy2013intriguing}
Christian Szegedy, Wojciech Zaremba, Ilya Sutskever, Joan Bruna, Dumitru Erhan,
  Ian Goodfellow, and Rob Fergus.
\newblock Intriguing properties of neural networks.
\newblock {\em arXiv preprint arXiv:1312.6199}, 2013.

\bibitem{tian2021analysis}
Qi Tian, Kun Kuang, Kelu Jiang, Fei Wu, and Yisen Wang.
\newblock Analysis and applications of class-wise robustness in adversarial
  training.
\newblock In {\em KDD}, 2021.

\bibitem{tsipras2018robustness}
Dimitris Tsipras, Shibani Santurkar, Logan Engstrom, Alexander Turner, and
  Aleksander Madry.
\newblock Robustness may be at odds with accuracy.
\newblock {\em arXiv preprint arXiv:1805.12152}, 2018.

\bibitem{wang2022self}
Hongjun Wang and Yisen Wang.
\newblock Self-ensemble adversarial training for improved robustness.
\newblock In {\em ICLR}, 2022.

\bibitem{wang2023simple}
Hongjun Wang and Yisen Wang.
\newblock Generalist: Decoupling natural and robust generalization.
\newblock In {\em CVPR}, 2023.

\bibitem{wang2021convergence}
Yisen Wang, Xingjun Ma, James Bailey, Jinfeng Yi, Bowen Zhou, and Quanquan Gu.
\newblock On the convergence and robustness of adversarial training.
\newblock In {\em ICML}, 2019.

\bibitem{wang2019improving}
Yisen Wang, Difan Zou, Jinfeng Yi, James Bailey, Xingjun Ma, and Quanquan Gu.
\newblock Improving adversarial robustness requires revisiting misclassified
  examples.
\newblock In {\em ICLR}, 2019.

\bibitem{wu2020adversarial}
Dongxian Wu, Shu-Tao Xia, and Yisen Wang.
\newblock Adversarial weight perturbation helps robust generalization.
\newblock In {\em NeurIPS}, 2020.

\bibitem{xie2019feature}
Cihang Xie, Yuxin Wu, Laurens van~der Maaten, Alan~L Yuille, and Kaiming He.
\newblock Feature denoising for improving adversarial robustness.
\newblock In {\em CVPR}, 2019.

\bibitem{xu2021robust}
Han Xu, Xiaorui Liu, Yaxin Li, Anil Jain, and Jiliang Tang.
\newblock To be robust or to be fair: Towards fairness in adversarial training.
\newblock In {\em ICML}, 2021.

\bibitem{zhang2019theoretically}
Hongyang Zhang, Yaodong Yu, Jiantao Jiao, Eric Xing, Laurent El~Ghaoui, and
  Michael Jordan.
\newblock Theoretically principled trade-off between robustness and accuracy.
\newblock In {\em ICML}, 2019.

\bibitem{zhang2020attacks}
Jingfeng Zhang, Xilie Xu, Bo Han, Gang Niu, Lizhen Cui, Masashi Sugiyama, and
  Mohan Kankanhalli.
\newblock Attacks which do not kill training make adversarial learning
  stronger.
\newblock In {\em ICML}, 2020.

\bibitem{GAIR}
Jingfeng Zhang, Jianing Zhu, Gang Niu, Bo Han, Masashi Sugiyama, and Mohan
  Kankanhalli.
\newblock Geometry-aware instance-reweighted adversarial training.
\newblock In {\em ICLR}, 2021.

\end{thebibliography}
}

\appendix
\newpage
\section{Details of the illustration Example}
\label{illustration Example}
In Sec.~\ref{sec:theoretical insight}, we use an illustration example to draw the theoretical results. Here we show the implementation detail of this toy example.

Note that for the case $d=1$, the data distribution is 
\begin{equation}
    x_1 = \begin{cases}
    + y,& \text{w.p.}\ p_y\\
    - y,& \text{w.p.}\ 1-p_{y}
    \end{cases}  \text{and} \quad
    x_2\overset{\text{i.i.d}}{\sim}\mathcal N(\eta y, \sigma^2).
\end{equation}

In this toy model, we select $p_{+1} = 0.85 > 0.7 = p_{-1}$ and $\eta=0.4$. The variance $\sigma^2$ is set to be 0.6 for better visualization in this toy model, and in the following theoretical analysis, we set $\sigma^2=1$ for simplicity. In Fig.~\ref{fig:toy model}(a), we randomly sample 100 pairs of $(x_1,x_2)$ for each class $y\in\{+1,-1\}$. In Fig.~\ref{fig:toy model}(b), the robustness is evaluated under perturbation bound $\epsilon=2\eta=0.8$, which is consistent to the evaluation in~\cite{tsipras2018robustness}.

\section{Proofs for Theorems in Sec.~\ref{sec:theoretical insight}}
\label{proof}
\subsection{Preliminaries}
We denote the distribution function and the  probability density function  of the \textit{normal distribution} $\mathcal N(0,1)$ as $\phi(x)$ and $\Phi(x)$:
\begin{equation}
    \begin{split}
        \Phi(x) = & \int_{-\infty}^x \frac{1}{\sqrt{2\pi}}e^{-\frac {x^2} {2}}{\mathrm d}t = \Pr.(\mathcal N(0,1) < x),\\
    \phi(x)   = & \frac{1}{\sqrt{2\pi}}e^{-\frac {x^2} {2}} = \Phi'(x).
    \end{split}
\end{equation}
Recall that the data distribution is 
\begin{equation}
\begin{split}
        &x_1 = \begin{cases}
    + y,& \text{w.p.}\ p_y,\\
    - y,& \text{w.p.}\ 1-p_{y},
    \end{cases} \\
        &x_2, \cdots, x_{d+1}\overset{\text{i.i.d}}{\sim}\mathcal N(\eta y, 1),
\end{split}
\end{equation}
where $1>p_{+1}>p_{-1}>\frac 1 2$.
First we calculate the clean accuracy $\mathcal A_y(f_w)$ and the robust accuracy $\mathcal R_y(f_w)$ for any class $y\in\{+1, -1\}$ and $w>0$. Also recall that the classifier 
\begin{equation}
    f_w=\text{sign}(x_1+\frac{x_2+\cdots+x_{d+1}}{w})
\end{equation}

Note that $w>0$, we have

\begin{equation}
\label{eq:A+1}
\begin{split}
    \mathcal A_{+1}(f_w) &= \Pr.(\text{sign}(f_w)=1)\\
    &= \Pr.(x_1 + \frac{x_2+\cdots +x_{d+1}}{w}>0)\\
    &= p_{+1}\cdot \Pr.(1 + \frac{x_2+\cdots +x_{d+1}}{w}>0) \\ 
    &+ (1-p_{+1})\cdot \Pr.(-1 + \frac{x_2+\cdots +x_{d+1}}{w}>0)\\
    &= p_{+1}\cdot \Pr.({x_2+\cdots +x_{d+1}}>-w) \\ 
    &+ (1-p_{+1})\cdot \Pr.(x_2+\cdots +x_{d+1}>w) \\
    &= p_{+1}\cdot \Pr.(\mathcal N(d\eta , d) > -w)\\ 
    &+ (1-p_{+1})\cdot \Pr.(\mathcal N(d\eta , d) > w) \\
    &= p_{+1} \cdot \Pr.(\mathcal N(0,d)> -d\eta-w)\\
    &+ (1-p_{+1}) \cdot \Pr. (\mathcal N(0,d) > -d\eta + w)\\
    &= p_{+1}\cdot \Pr. (\mathcal N(0,1) < \frac{d\eta + w}{\sqrt d})\\
    &+ (1-p_{+1})\cdot \Pr. (\mathcal N(0,1) < \frac{d\eta - w}{\sqrt d})\\
    & = p_{+1}\Phi(\frac{d\eta + w}{\sqrt d}) + (1-p_{+1}) \Phi(\frac{d\eta - w}{\sqrt d}).
\end{split}
\end{equation}
Similarly, we have 
\begin{equation}
\label{eq:A-1}
    \mathcal A_{-1}(f_w) = p_{-1}\Phi(\frac{d\eta + w}{\sqrt d}) + (1-p_{-1}) \Phi(\frac{d\eta - w}{\sqrt d}).
\end{equation}
For the robustness, following the evaluation in the original model~\cite{tsipras2018robustness}, we evaluate the  robustness $\mathcal R_y$ under $l_\infty$-norm perturbation bound $\epsilon=2\eta<1$. Consider the distribution of adversarial examples $\hat x = (\hat x_1, \hat x_2, \cdots, \hat x_{d+1})$. Since we restrict the robust feature $x_1\in\{-1, +1\}$ and $\epsilon < 1$, we have $\hat x_1 = x_1$. For the non-robust features $x_i\sim\mathcal N(\eta y, 1)$, the corresponding adversarial example has $\hat x_i\sim \mathcal N(-\eta y, 1)$ under the perturbation bound $\epsilon=2\eta$. Therefore, the distribution of adversarial examples is
\begin{equation}
\label{eval_adv_distribution}
    \hat x_1 = \begin{cases}
    + y,& \text{w.p.}\ p_y\\
    - y,& \text{w.p.}\ 1-p_{y}
    \end{cases}  \text{and} \quad
    \hat x_2, \cdots, \hat x_{d+1}\overset{\text{i.i.d}}{\sim}\mathcal N(-\eta y, 1).
\end{equation}
By simply replacing $\eta$ with $-\eta$ in derivative process of (\ref{eq:A+1}), for any $w>0$, we have
\begin{equation}
\label{R}
\begin{split}
    \mathcal R_{+1}(f_w) &= p_{+1}\Phi(\frac{-d\eta + w}{\sqrt d}) + (1-p_{+1}) \Phi(\frac{-d\eta - w}{\sqrt d}),\\
    \mathcal R_{-1}(f_w) &= p_{-1}\Phi(\frac{-d\eta + w}{\sqrt d}) + (1-p_{-1}) \Phi(\frac{-d\eta - w}{\sqrt d}).
\end{split}
\end{equation}

\subsection{Proof of Theorem~\ref{theorem:difficulty}}
The theorem~\ref{theorem:difficulty} shows the class $y=-1$ is intrinsically difficult to learn than class $y=+1$:

\noindent\textbf{Theorem 1}
For any $w>0$ and the classifier $f_w=\text{sign}(x_1+\frac{x_2+\cdots+x_{d+1}}{w})$, we have $\mathcal A_{+1}(f_w) > \mathcal A_{-1} (f_w)$ and  $\mathcal R_{+1}(f_w)>R_{-1}(f_w)$.

\noindent\textit{Proof.} Note that $p_{+1} > p_{-1}$, and $\Phi(\frac{d\eta + w}{\sqrt d})> \Phi(\frac{d\eta - w}{\sqrt d})$, we have
\begin{equation}
\begin{split}
\mathcal{A}_{+1}(f_w)&=p_{+1}\Phi(\frac{d\eta + w}{\sqrt d}) + (1-p_{+1}) \Phi(\frac{d\eta - w}{\sqrt d})  \\
&= p_{+1}(\Phi(\frac{d\eta + w}{\sqrt d})- \Phi(\frac{d\eta - w}{\sqrt d})) + \Phi(\frac{d\eta - w}{\sqrt d}) \\
&>p_{-1}(\Phi(\frac{d\eta + w}{\sqrt d})- \Phi(\frac{d\eta - w}{\sqrt d})) + \Phi(\frac{d\eta - w}{\sqrt d}) \\
&= \mathcal{A}_{-1}(f_w).
\end{split}
\end{equation}

\subsection{Proof of Theorem~\ref{theorem:eps}}
The theorem~\ref{theorem:eps} shows the relation between the parameter $w$ and the attack strength (perturbation bound $\epsilon$) in adversarial training:

\noindent\textbf{Theorem 2} For any $0\le\epsilon\le\eta$, the optimal parameter $w$ for adversarial training with perturbation bound $\epsilon$ is monotone increasing at $\epsilon$.

\noindent\textit{Proof.} Similar to the adversarial example distribution analysis (\ref{eval_adv_distribution}), under the perturbation bound $\epsilon$, the data distribution of the crafted adversarial example for training is 
\begin{equation}
\label{train data distribution}
\begin{split}
    &\tilde x_1 = \begin{cases}
    + y,& \text{w.p.}\ p_y\\
    - y,& \text{w.p.}\ 1-p_{y}
    \end{cases},\\
    &\tilde x_2, \cdots, \tilde x_{d+1}\overset{\text{i.i.d}}{\sim}\mathcal N((\eta-\epsilon) y, 1).
\end{split}
\end{equation}
We use $\tilde {\mathcal A}(f_w)$, $\tilde {\mathcal A}_y(f_w)$ to denote the overall and class-wise \textit{train accuracy} of the classifier $f_w$ on training data distribution (\ref{train data distribution}).
Let $p = p_{+1}+p{-1}$.
Then the overall train accuracy of $f_w$ is 
\begin{equation}
\begin{split}
   & \tilde{\mathcal A}(f_w)= \frac{1}{2}(\tilde{\mathcal A}_{+1}(f_w) + \tilde{\mathcal A}_{-1}(f_w)) \\
   & =\frac{1}{2}(p_{+1}\Phi(\frac{d(\eta-\epsilon) + w}{\sqrt d}) + (1-p_{+1}) \Phi(\frac{d(\eta-\epsilon) - w}{\sqrt d})\\
   & + p_{-1}\Phi(\frac{d(\eta-\epsilon) + w}{\sqrt d}) + (1-p_{-1}) \Phi(\frac{d(\eta-\epsilon) - w}{\sqrt d})
    )\\
    & = \frac{1}{2}(p\Phi(\frac{d(\eta-\epsilon) + w}{\sqrt d}) + (2-p)\Phi(\frac{d(\eta-\epsilon) - w}{\sqrt d})).
\end{split}
\end{equation}

Now we calculate the best parameter $w$ for $\tilde{\mathcal A}(f_w)$. Note that $\Phi'(x)=\phi(x)$, we have

\begin{equation}
\begin{split}
    & \frac{\partial \tilde{\mathcal A}(f_w)}{\partial w}=\frac {1}{2\sqrt d} (p\phi(\frac{d(\eta-\epsilon) + w}{\sqrt d}) - (2-p)\phi(\frac{d(\eta-\epsilon) - w}{\sqrt d}))\\
   &  = \frac {1}{2\sqrt{2\pi d}}\{ p\exp[-\frac 1 2(\frac{d(\eta-\epsilon) + w}{\sqrt d})^2]\\& -(2-p)\exp[-\frac 12(\frac{d(\eta-\epsilon)-w}{\sqrt{d}})^2] \}
\end{split}
\end{equation}
Therefore, $\frac{\partial \tilde{\mathcal A}(f_w)}{\partial w}>0$ is equivalent to
\begin{equation}
\begin{split}
&p\exp[-\frac 1 2(\frac{d(\eta-\epsilon) + w}{\sqrt d})^2]> (2-p)\exp[-\frac 12(\frac{d(\eta-\epsilon)-w}{\sqrt{d}})^2]\\
&\iff \exp[-\frac 1 2((\frac{d(\eta-\epsilon) + w}{\sqrt d})^2-(\frac{d(\eta-\epsilon)-w}{\sqrt{d}})^2)]>\frac {2-p}{p}\\
&\iff \exp[-\frac 1 {2d} \cdot (4d(\eta-\epsilon)w)]>\frac {2-p}{p}\\
&\iff \exp[-2(\eta-\epsilon)w]>\frac {2-p}{p}\\
&\iff -2(\eta-\epsilon)w > \ln(\frac {2-p}{p})\\
&\iff w < \frac{1}{2(\eta-\epsilon)}\ln(\frac{p}{2-p}):=\hat w_\epsilon.
\end{split}
\end{equation}
Recall that we assume $p_{+1}, p_{-1}>\frac 1 2$, thus $p=p_{+1}+p_{-1}>1$ and $\frac p {2-p} > 1$. 
Therefore, $\frac{\partial \tilde{\mathcal A}(f_w)}{\partial w}>0$ when $w<\hat w_\epsilon$, and $\frac{\partial \tilde{\mathcal A}(f_w)}{\partial w}<0$ when $w>\hat w_\epsilon$. We can conclude that $f_w$ obtains the optimal parameter $w$, \textit{i.e.}, $w$ achieves the highest train accuracy,  when $w=\hat w_\epsilon = \frac{1}{2(\eta-\epsilon)}\ln(\frac{p}{2-p})$, which is monotone increasing at $\epsilon$.

\subsection{Proof of Theorem~\ref{theorem:best}}
Theorem~\ref{theorem:best} shows the clean accuracy of the hard class $y=-1$ drops earlier than class $y=+1$ as  the attack strength increases:

\noindent\textbf{Theorem 3} Let $w_y^* = \arg\max\limits_w \mathcal A_y(f_w)$ be the parameter for the best clean accuracy of class $y$, then $w_{+1}^*>w_{-1}^*$.

\noindent\textit{Proof.} 
As calculated in (\ref{eq:A+1}) and (\ref{eq:A-1}), we have $\mathcal A_y(f_w)  = p_y\Phi(\frac{d\eta + w}{\sqrt d}) + (1-p_y) \Phi(\frac{d\eta - w}{\sqrt d})$ and 
\begin{equation}
\begin{split}
    &\frac{\partial {\mathcal A}(f_w)}{\partial w} = \frac{1}{\sqrt d}(p_y\phi(\frac{d\eta + w}{\sqrt d}) - (1-p_y) \phi(\frac{d\eta - w}{\sqrt d})).
    \end{split}
\end{equation}

Therefore, $\frac{\partial {\mathcal A}(f_w)}{\partial w}>0$ is equivalent to
\begin{equation}
\begin{split}
    &\exp\{-\frac 1 2 [(\frac{d\eta + w}{\sqrt d})^2 - (\frac{d\eta - w}{\sqrt d}) ^2]\} > \frac{1-p_y}{p_y}\\
    &\iff \exp\{-2\eta w\}>\frac{1-p_y}{p_y}\\
    & \iff -2\eta w > \ln (\frac{1-p_y}{p_y})\\
    & \iff w < \frac{1}{2\eta}\ln(\frac{p_y}{1-p_y}).
\end{split}
\end{equation}
Similar to the proof of Theorem 2, we have $w^*_y = \arg\max \mathcal A_y(f_w)=\frac{1}{2\eta}\ln(\frac{p_y}{1-p_y})$. Since $1>p_{+1}>p_{-1}>\frac 1 2$, we have $\frac{p_{+1}}{1-p_{+1}}>\frac{p_{-1}}{1-p_{-1}}>1$ and hence $w^*_{+1}>w^*_{-1}$.

\subsection{Proof of Theorem~\ref{theorem:compare}}
Theorem 4 shows how strong attack in adversarial training hurts the hard class $y=-1$:

\noindent\textbf{Theorem 4} Suppose $\Delta_w>0$, then for $\forall w>w_{+1}^*
$, $\mathcal A_{-1}(f_{w+\Delta_w}) - \mathcal A_{-1}(f_w) < \mathcal A_{+1}(f_{w+\Delta_w}) - \mathcal A_{+1}(f_w)<0$, and for $\forall w>0$, $0<\mathcal R_{-1}(f_{w+\Delta_w}) - \mathcal R_{-1}(f_w) < \mathcal R_{+1}(f_{w+\Delta_w}) - \mathcal R_{+1}(f_w)$. 

\noindent \textit{Proof.} First we prove for $u>w^*_{+1}$,
\begin{equation}
    \mathcal A_{-1}(f_{w+\Delta_w}) - \mathcal A_{-1}(f_w) < \mathcal A_{+1}(f_{w+\Delta_w}) - \mathcal A_{+1}(f_w)<0.
\end{equation}
Since we have
\begin{equation}
\begin{split}
    \mathcal A_{y}(f_{w+\Delta_w}) - \mathcal A_{y}(f_w)=\int_{w}^{w+\Delta w} \frac{\partial {\mathcal A_y}(f_u)}{\partial u} \mathrm{d} u,
\end{split}
\end{equation}
It's suffice to show that 
\begin{equation}
\frac{\partial {\mathcal A_{-1}}(f_u)}{\partial u}<\frac{\partial {\mathcal A_{+1}}(f_u)}{\partial u}<0,\quad \forall u>w^*_{+1}.
\end{equation}

Recall that in the proof of Theorem 3, we have shown \begin{equation}
\begin{split}
&\frac{\partial {\mathcal A}(f_u)}{\partial w} = \frac{1}{\sqrt d}(p_y\phi(\frac{d\eta + w}{\sqrt d}) - (1-p_y) \phi(\frac{d\eta - w}{\sqrt d}))\\
& = \frac{1}{\sqrt d} \{p_y[\phi(\frac{d\eta + w}{\sqrt d})+ \phi(\frac{d\eta - w}{\sqrt d})]-\phi(\frac{d\eta - w}{\sqrt d})\}.
\end{split}
\end{equation}
Therefore, since $p_{-1}<p_{+1}$ and $\phi(\frac{d\eta + w}{\sqrt d})+ \phi(\frac{d\eta - w}{\sqrt d}) > 0$, we have 
\begin{equation}
    \frac{\partial {\mathcal A_{-1}}(f_u)}{\partial u}<\frac{\partial {\mathcal A_{+1}}(f_u)}{\partial u}.
\end{equation}
Further, since $u>w^{*}_{+1}$, we have $\frac{\partial {\mathcal A_{+1}}(f_u)}{\partial u}<0$ as shown in the proof of Theorem 3.

Next, we prove that for $\forall w>0$,
\begin{equation}
    0<\mathcal R_{-1}(f_{w+\Delta_w}) - \mathcal R_{-1}(f_w) < \mathcal R_{+1}(f_{w+\Delta_w}) - \mathcal R_{+1}(f_w).
\end{equation}
Similarly, it suffice to show
\begin{equation}
    0<\frac{\partial {\mathcal R_{-1}}(f_u)}{\partial u}<\frac{\partial {\mathcal R_{+1}}(f_u)}{\partial u},\quad \forall u>0.
\end{equation}

Recall the expression (\ref{R}), we have
\begin{equation}
    \mathcal R_y =  p_{y}\Phi(\frac{-d\eta + w}{\sqrt d}) + (1-p_y) \Phi(\frac{-d\eta - w}{\sqrt d}),\\
\end{equation}
hence
\begin{equation}
\begin{split}
    &\frac{\partial \mathcal R_y(f_w)}{\partial w} = \frac{1}{\sqrt d}
    \{p_y\phi(\frac{-d\eta + w}{\sqrt d}) - (1-p_y)\phi(\frac{-d\eta - w}{\sqrt d})\}\\
    &=\frac{1}{\sqrt d}\{ p_y[\phi(\frac{-d\eta + w}{\sqrt d})+\phi(\frac{-d\eta - w}{\sqrt d})] - \phi(\frac{-d\eta - w}{\sqrt d})\}\\
\end{split}
\end{equation}
Since $p_{+1}>p_{-1}$ and $\phi(\frac{-d\eta + w}{\sqrt d})+\phi(\frac{-d\eta - w}{\sqrt d})>0$, we have 
\begin{equation}
    \frac{\partial {\mathcal R_{-1}}(f_u)}{\partial u}<\frac{\partial {\mathcal R_{+1}}(f_u)}{\partial u}.
\end{equation}
Finally, as $d, \eta, w>0$, we have $(\frac{-d\eta + w}{\sqrt d})^2<(\frac{-d\eta - w}{\sqrt d})^2$ by comparing their absolute value. This indicates $\phi(\frac{-d\eta + w}{\sqrt d})>\phi(\frac{-d\eta - w}{\sqrt d})$. Also note that $p_{-1}>\frac 1 2$ and $p_{-1} > (1-p_{-1})$, we have 
\begin{equation}
 \frac{1}{\sqrt d}
    \{p_{-1}\phi(\frac{-d\eta + w}{\sqrt d}) - (1-p_{-1})\phi(\frac{-d\eta - w}{\sqrt d})\}>0,
\end{equation}
which completes our proof.

\section{More Experiments}
\label{More experiment}
Here we present additional experimental results.

\subsection{Experiment on Tiny-ImageNet}
\label{C1}
Besides CIFAR-10, we additionally compare CFA with baseline+EMA on \textbf{Tiny-ImageNet} with ResNet-18 under $\ell_\infty$-norm bound $\epsilon=4/255$.
Since the worst class robustness is extremely low and there are only 50 images for each class in the test set, we report the average of the worst-20\% class robustness. The threshold of FAWA is also set on the average robustness of these classes  on validation set.
The results in Table~\ref{tab:ti} shows that our CFA framework still outperforms baseline+EMA on Tiny-ImageNet.

\subsection{Class-wise robustness comparison}
\label{C2}
\begin{figure}[!t]
    \centering
    \includegraphics[width=0.4\textwidth]{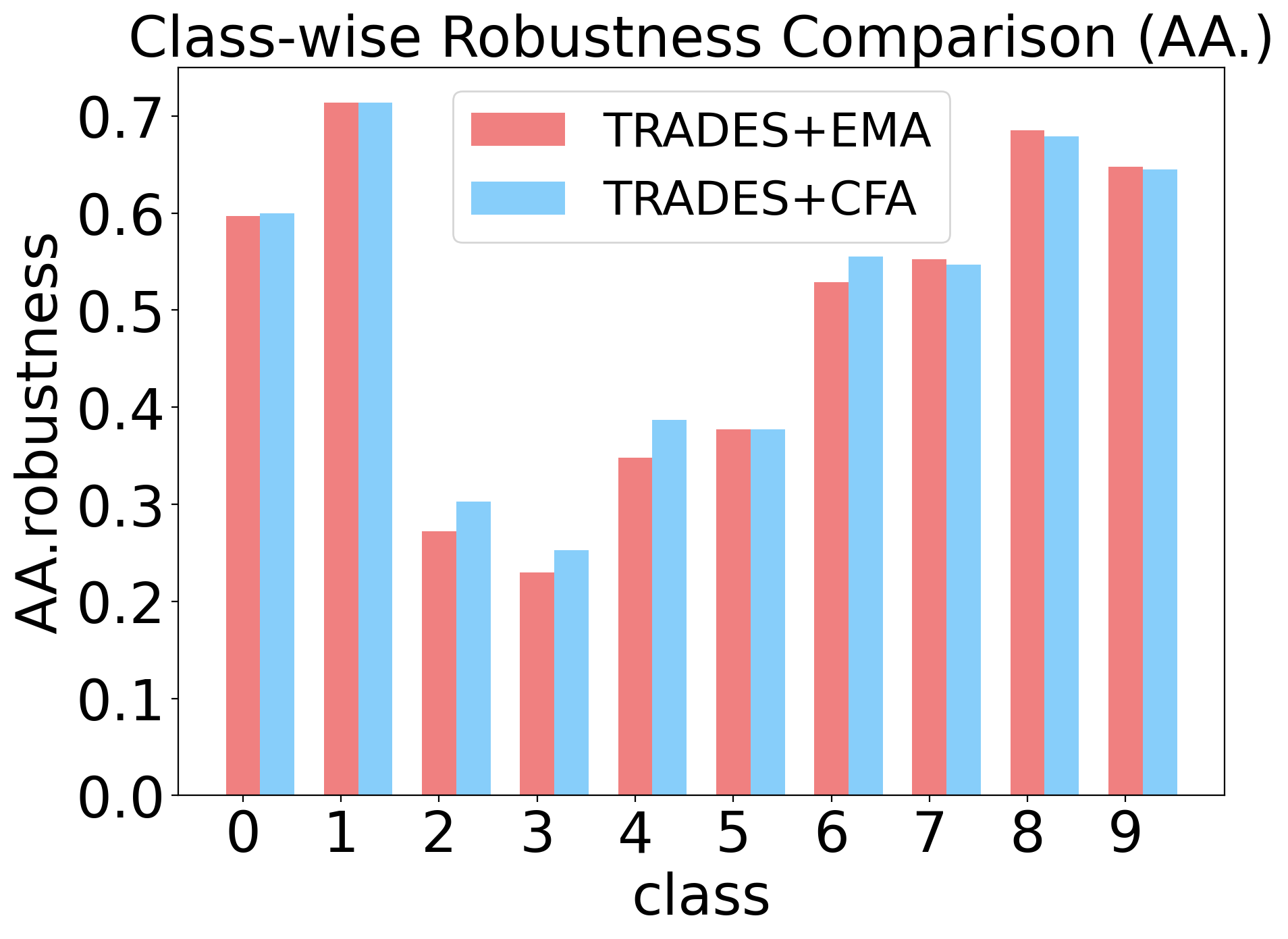}
    \caption{Class-wise robustness comparison between TRADES+EMA and TRADES+CFA on CIFAR-10 dataset at the best checkpoint. Robustness evaluated under AutoAttack.}
    \label{fig:class-wise}
\end{figure}

To evaluate class-wise robustness, we present a comparison between CFA and EMA on CIFAR-10, as shown in Fig.~\ref{fig:class-wise}. The results show that
CFA significantly outperforms EMA on classes \{2,3,4,6\}, while slightly dropping on classes \{7,8\}.
Compared with the improvements, the decreases are very slight. 
Moreover, the variance of class-wise robustness, which measures differences between classes,  is also lower for CFA (0.15) compared to EMA (0.17). This indicates that CFA indeed reduces the difference among class-wise robustness and improves the fairness without harming other classes.

\subsection{Selection of $\lambda_2$}
\label{C3}
\begin{figure}[h]
    \centering
    \begin{tabular}{cc}
        \includegraphics[width=0.22\textwidth]{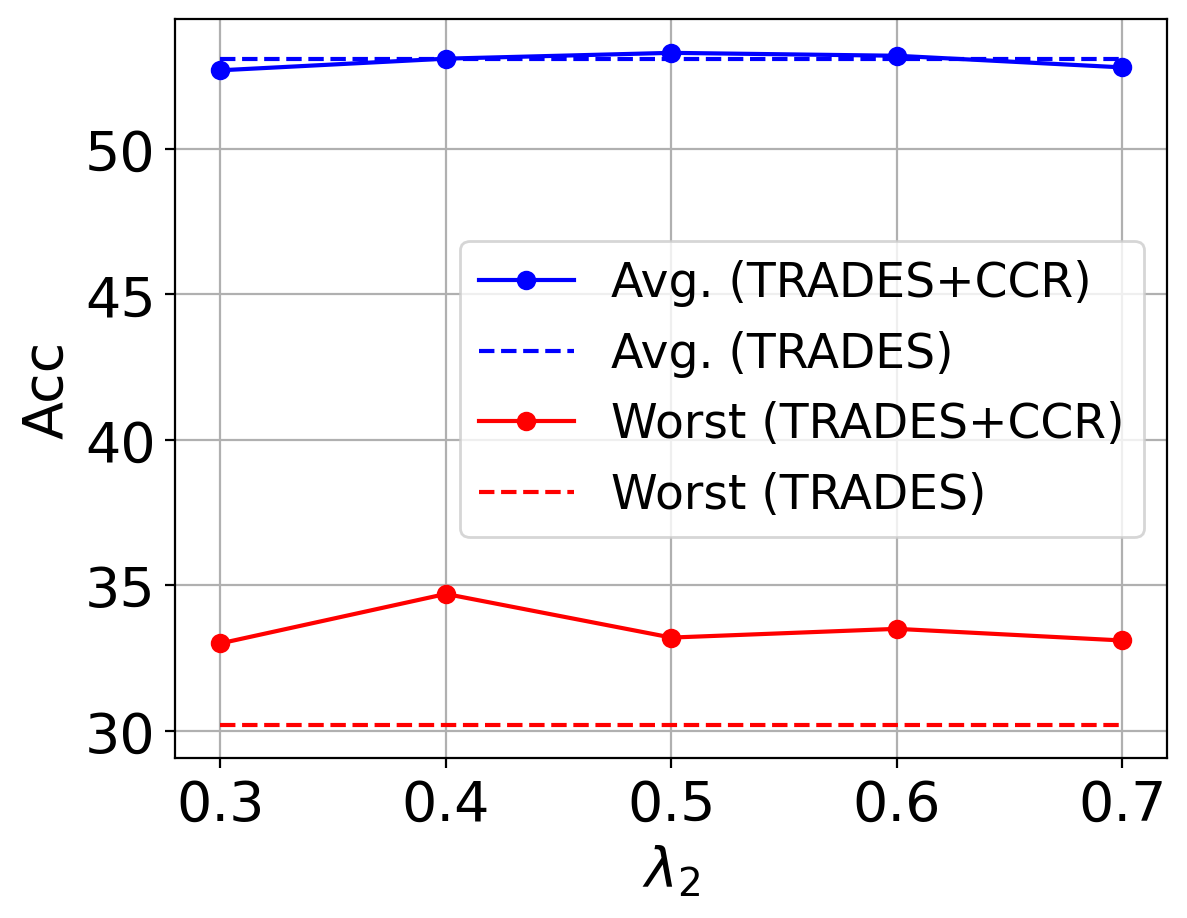}  &  \includegraphics[width=0.22\textwidth]{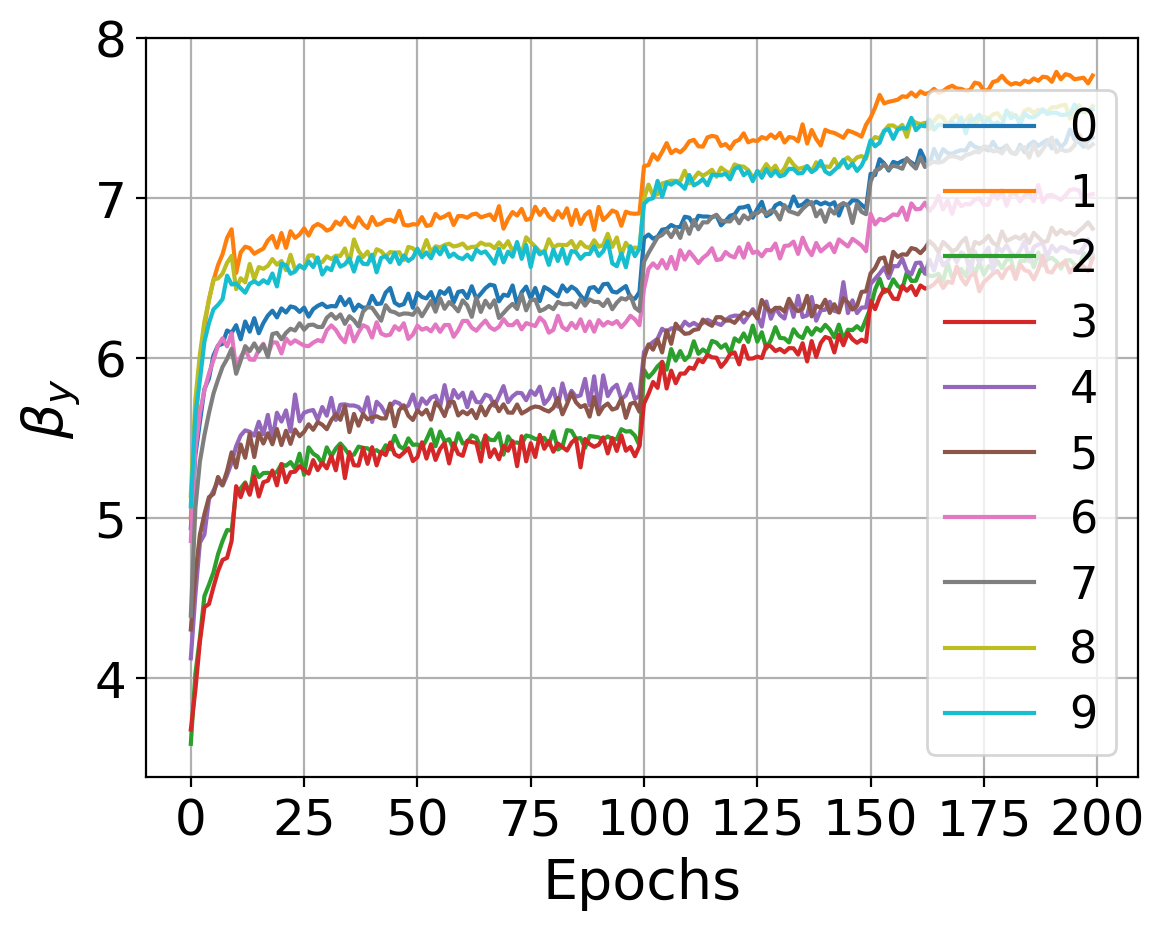}
         \\
        (a) & (b)
    \end{tabular}
    \caption{Analysis on the base regularization budget $\lambda_2$. (a): Average and the worst class robustness of models trained with different $\lambda_2$ (solid) and vanilla TRADES (dotted). (b): Class-wise calibrated regularization $\beta_y$ in the training phase of $\lambda_2=0.4$.}
    \label{fig:ccr analysis}
\end{figure}

Following our analysis on the selection of the perturbation budget $\lambda_1$ for AT+CCM in Sec.~\ref{ablation}, we conduct a similar analysis on the influence of regularization budget $\lambda_2$ for TRADES + CCM  + CCR in Fig.~\ref{fig:ccr analysis}.

In Fig.~\ref{fig:ccr analysis}(a), we compare the selection of $\lambda_2$ from 0.3 to 0.7. The robustness is evaluated under PGD-10.
The base perturbation budget $\lambda_1$ of CCM is still selected as 0.3. Comparing to vanilla TRADES, our TRADES+CCM+CCR outperforms in the worst class robustness significantly, and the overall robustness is marginal higher than TRADES for $\lambda_2=0.4$, 0.5 and 0.6.

Fig.~\ref{fig:ccr analysis}(b) shows the $\beta_y$ used in the case $\lambda_2=0.4$. We can see that the hard classes use $\beta_y\approx6$, while the easy classes use higher $\beta_y$. This is consistent to our analysis on class-wise robustness under different regularization $\beta$ in Sec~\ref{regularizations}.

\begin{table}[!t]
    \centering
    \small
    \caption{Overall comparison of experiment on Tiny-ImageNet.}
    \resizebox{\linewidth}{!}{
    \begin{tabular}{l|cc|cc}
    \toprule[2pt]
        \textbf{Tiny-ImageNet} & \multicolumn{2}{c|}{Best (Avg. / Worst-20\%)} & \multicolumn{2}{c}{Last (Avg. / Worst-20\%)} 
        \\
        \textbf{Method} & Clean   & AA. & Clean & AA.
        \\ \midrule[1pt]
        AT & 41.1/\textbf{16.4}
        & 20.2/3.6 & 39.4/17.3 & 14.9/1.6 \\
        AT + EMA & 
40.7/14.7 & 21.8/4.2 & 41.6/19.8 & 17.4/3.9
        \\
AT + CFA & 
\textbf{41.2}/{16.2} & \textbf{22.4}/\textbf{5.2} & \textbf{42.3}/\textbf{20.0} & \textbf{19.9}/\textbf{4.8}
        \\
        \midrule[1pt]
        TRADES & \textbf{43.2}/18.4
        & 20.9/3.7 & 42.5/18.7 & 18.8/3.4
        \\
TRADES + EMA &
41.2/19.5  & 21.6/4.1 & \textbf{43.3}/\textbf{19.8} & 19.9/3.8
        \\
TRADES + CFA&
{41.7}/\textbf{20.0} & \textbf{22.3}/\textbf{5.5} & 42.4/19.6 & \textbf{21.2}/\textbf{5.2}
        
        \\         
        \midrule[1pt]
        FAT
        & 43.6/19.2 & 19.2/2.6 & 39.7/17.8 & 14.3/1.7
        \\
        FAT + EMA &
43.4/18.6 & 21.0/4.1 & 42.9/19.9 & 17.0/2.6
        
        \\
        FAT + CFA & 
\textbf{43.7}/\textbf{19.6} & \textbf{21.6}/\textbf{4.9} & \textbf{43.6}/\textbf{21.3} & \textbf{19.1}/\textbf{3.4}
        
        \\
        \bottomrule[2pt]
        \end{tabular}}

    \label{tab:ti}
\end{table}

\end{document}